\documentclass[11pt]{article}

\usepackage[final]{acl}

\usepackage{times}
\usepackage{latexsym}

\usepackage[T1]{fontenc}

\usepackage[utf8]{inputenc}

\usepackage{microtype}

\usepackage{inconsolata}

\usepackage{graphicx}
\usepackage{subcaption}
\usepackage{amsmath}
\usepackage{enumitem} 
\usepackage{booktabs}
\usepackage{tikz}
\usetikzlibrary{arrows.meta, shapes.geometric}
\usepackage{multirow}
\usepackage{amssymb} 
\usetikzlibrary{positioning}
\usepackage{float}
\usepackage{xcolor}

\title{PolERo: Studying Political Evasion in Romanian}

\author{Gabriel Ștefan \and Sergiu Nisioi\thanks{Corresponding authors.} \\
     Laboratory for Cybernetic Research and Good Life \\
     Human Language Technologies Research Center \\
     Faculty of Mathematics and Computer Science \\ University of Bucharest \\ \texttt{gabrielstefan04@gmail.com} \\ \texttt{sergiu.nisioi@unibuc.ro}}

\begin{document}
\maketitle
\begin{abstract} 
Political evasion refers to responses that engage with a question while withholding the requested information. Recent NLP work frames political evasion as a classification task using a two-level taxonomy of response clarity and fine-grained evasion strategies. Existing work on response clarity and evasion classification is limited to English, leaving open whether the taxonomy and model behavior transfer across languages and political contexts.  We introduce \textbf{PolERo}, a dataset of $3{,}574$ human-annotated question-answer pairs extracted from official transcripts of five Romanian presidents. We evaluate multiple classification approaches on both datasets under matched conditions, including TF-IDF baselines, fine-tuned encoder models, a proposed sliding-window encoder, and zero/few-shot LLM prompting. We study cross-lingual transfer through joint bilingual training and machine-translation-based data augmentation. Our results indicate that fine-tuned encoders are competitive, cross-lingual transfer is asymmetric, and ambivalent evasion categories involving pragmatic cues remain the main challenge across all model families.
\end{abstract}

\section{Introduction}

Political interviews often contain responses that are relevant to the interviewer's questions while at the same time remaining non-responsive. Such responses may involve implicit answers, partial replies, topic shifts \cite{topicshift}, deflections, or strategic ambiguity \cite{bach2025let}, making it difficult to determine whether the requested information was actually provided. In political science and discourse analysis, these phenomena are commonly studied under the notions of \textbf{equivocation} and \textbf{evasion} \citep{bavelas1988political, bull1994identifying, clayman2001answers, bull2003, rasiah2010, romaniuk2013pursuing, bull2020can}.

Beyond political communication, response clarity is also relevant to broader NLP problems including dialogue understanding, question answering, and implicitly for creating better user-facing technologies. Despite recent progress in political discourse processing \cite{glavas-etal-2019-computational,huguet-cabot-etal-2020-pragmatics,politicalnlp-2024-natural}, response clarity remains largely unexplored in NLP. Existing work has focused primarily on question answerability \citep{squad, ambigqa}, responder intent \citep{ferracane-etal-2021-answer}, or broader discourse phenomena in political speech \cite{subramanian-etal-2019-target, reinig-etal-2024-politics, verma-etal-2026-predicting}, while leaving open the problem of determining whether a response adequately addresses a question. Unlike traditional QA settings \cite{squad, ambigqa, clark-etal-2020-tydi}, response clarity requires reasoning about pragmatic alignment, modeling information sufficiency \cite{jain-garimella-2026-knowing}, and discourse structure \cite{mann-1984-discourse}, extending beyond standard answerability formulations.

Recently, \citet{thomas2024} introduced the first NLP dataset and taxonomy of 3,400 question-answer (QA) pairs from the U.S. presidential interviews. The dataset served as a reference for SemEval-2026 Task 6: CLARITY \citep{thomas2026clarity} a task that aims to detect and classify response ambiguity in political discourse.  Evidence for response clarity classification currently remains restricted to U.S. English political interviews. Furthermore, response clarity is not a purely linguistic phenomenon; it is shaped by institutional context, interaction format, and political communication practices. Currently, no resource targets response clarity or evasion classification in Romanian political speech. To the best of our knowledge, such resources are also absent for non-English languages more broadly.

To address this, we introduce PolERo (Political Evasion in Romanian), a dataset of 3,574 human-annotated question-answer pairs extracted from official transcripts covering five Romanian presidents. PolERo adopts the exact annotation protocol and two-level taxonomy proposed by \citet{thomas2024}, enabling controlled comparison between English and Romanian political interviews. 

Romanian is an Eastern Romance language with historically limited NLP resources. Recent efforts have significantly expanded this landscape, introducing general evaluation benchmarks and foundational corpora \citep{dumitrescu-avram-2020-introducing, liro2021, medqaro}, alongside Romanian-adapted language models \citep{masala-etal-2020-robert, dumitrescu-etal-2020-birth, dima-etal-2024-roqllama, masala2024openllmrotechnicalreport}. Furthermore, Romanian political communication has attracted sustained interest in discourse studies and political linguistics with a focus on political opinion mining \citep{gifu2013towards, vasilescu2024insights} and hate-speech discourse \cite{gheorghiu2022hate,dinu-etal-2025-antisemro}.

Nevertheless, despite this progress, no prior datasets or analyses target response clarity or evasion classification in any type of speech. This makes PolERo, to the best of our knowledge, the first such resource for a non-English language. This paper makes several contributions:
\begin{itemize}[itemsep=2pt, topsep=3pt]
    \item The release of PolERo,\footnote{The dataset, code, and prompts are released under the CC BY 4.0 License and are available at \url{https://github.com/gabriel-stefan/polero}.} a human-annotated dataset for response clarity and evasion classification in a Romanian political corpus.
    \item Multiple classification results for both PolERo and the English dataset of \citet{thomas2024} under matched conditions, covering encoder models, a multi-head encoder for long text classification, and different LLM prompting strategies.
    \item A cross-lingual transfer analysis examining the consistency of evasion patterns and model behavior across the two corpora, considering low-data scenarios and machine-translation-based data augmentation.
\end{itemize}

\section{Dataset and Annotation}

\subsection{Data Collection and Preprocessing}

We collect official transcripts from the Romanian presidential websites,\footnote{\url{https://www.presidency.ro} and \url{https://archive.presidency.ro}} covering official press conferences, interviews, and declarations from five Romanian presidents: Ion Iliescu, Traian B\u{a}sescu, Klaus Iohannis, Ilie Bolojan, and Nicușor Dan. The corpus spans 25 years between 2001--2026.

In such contexts, it is common for a single interviewer question to contain multiple sub-questions. The initial pool of extracted data contained 6{,}234 raw question-response candidates, comprising 8{,}233 sub-questions in total. For the current task, we enforce a one-to-one question-response mapping, and we use GPT-5.4 to flag the set of instances where more than one question is asked in a single instance. As such, we remove 2{,}053 multi-part entries and 607 entries containing no extractable question. During the annotation process, evaluators were instructed to flag any remaining double-barreled questions in the labeling platform (see the appendix \ref{app:annotation:platform}). No such instances were reported, indicating that no remaining multi-part questions were present in the retained corpus. The retained corpus consists of \textbf{3{,}574} single-part QA pairs. We divide the dataset into a training set of $3{,}278$ and a test set of 296 examples.

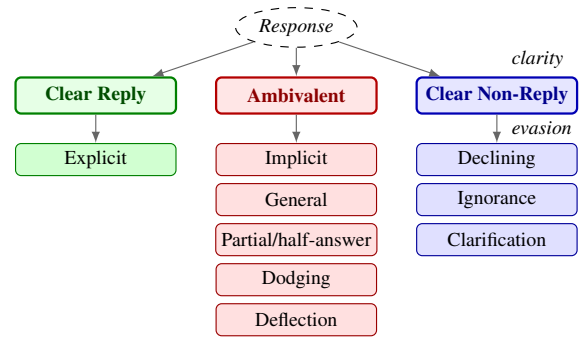
\begin{figure}[!htb]
\centering
\resizebox{\columnwidth}{!}{
\begin{tikzpicture}[
  font=\scriptsize,
  l1cr/.style={draw=green!50!black, thick, rounded corners=2pt, inner sep=2pt, minimum width=2.0cm, minimum height=0.45cm, font=\scriptsize\bfseries, fill=green!10, text=green!30!black},
  l1am/.style={draw=red!70!black,   thick, rounded corners=2pt, inner sep=2pt, minimum width=2.0cm, minimum height=0.45cm, font=\scriptsize\bfseries, fill=red!10,   text=red!55!black},
  l1cn/.style={draw=blue!70!black,  thick, rounded corners=2pt, inner sep=2pt, minimum width=2.0cm, minimum height=0.45cm, font=\scriptsize\bfseries, fill=blue!10,  text=blue!55!black},
  l2cr/.style={draw=green!50!black, rounded corners=2pt, inner sep=2pt, minimum width=2.0cm, minimum height=0.40cm, font=\scriptsize, fill=green!18},
  l2am/.style={draw=red!60!black,   rounded corners=2pt, inner sep=2pt, minimum width=2.0cm, minimum height=0.40cm, font=\scriptsize, fill=red!12},
  l2cn/.style={draw=blue!60!black,  rounded corners=2pt, inner sep=2pt, minimum width=2.0cm, minimum height=0.40cm, font=\scriptsize, fill=blue!12},
  root/.style={draw, dashed, ellipse, inner sep=2pt, font=\scriptsize\itshape},
  arr/.style={-latex, thin, draw=black!60}
]

\node[root] (R) at (0, 0) {Response};

\node[l1cr] (CR)  at (-2.5, -0.85) {Clear Reply};
\node[l1am] (AMB) at (0,    -0.85) {Ambivalent};
\node[l1cn] (CNR) at (2.5,  -0.85) {Clear Non-Reply};

\draw[arr] (R) -- (CR);
\draw[arr] (R) -- (AMB);
\draw[arr] (R) -- (CNR);

\node[l2cr] (EX)   at (-2.5, -1.65) {Explicit};

\node[l2am] (IMP)  at (0, -1.65) {Implicit};
\node[l2am] (GEN)  at (0, -2.15) {General};
\node[l2am] (PAR)  at (0, -2.65) {Partial/half-answer};
\node[l2am] (DOD)  at (0, -3.15) {Dodging};
\node[l2am] (DEF)  at (0, -3.65) {Deflection};

\node[l2cn] (DECL) at (2.5, -1.65) {Declining};
\node[l2cn] (IGN)  at (2.5, -2.15) {Ignorance};
\node[l2cn] (CLAR) at (2.5, -2.65) {Clarification};

\draw[arr] (CR.south)  -- (EX.north);
\draw[arr] (AMB.south) -- (IMP.north);
\draw[arr] (CNR.south) -- (DECL.north);

\node[font=\scriptsize\itshape, anchor=west] at (2.55, -0.4) {clarity};
\node[font=\scriptsize\itshape, anchor=west] at (2.55, -1.25) {evasion};

\end{tikzpicture}
}
\caption{Clarity-evasion two-level taxonomy mapping.}
\label{fig:taxonomy_mapping}
\end{figure}

\subsection{Taxonomy}
\label{sec:taxonomy}

PolERo adopts the two-level response clarity taxonomy of \citet{thomas2024}, focusing on the distinction between clarity and ambiguity in political responses (see \autoref{fig:taxonomy_mapping}). The first level assigns each response one of three clarity labels: \textit{Clear Reply} --- the information is given in the requested form, \textit{Ambivalent} --- a valid answer is given but admits multiple interpretations, or \textit{Clear Non-Reply} --- the speaker openly refuses to share the information. The second level divides these classes into nine fine-grained evasion categories. \textit{Clear Reply} contains only \textit{Explicit} information stated in the requested form. 

The \textit{Ambivalent} category includes cases where: 1. \textit{Implicit} information is conveyed without being explicitly stated; 2. \textit{General} information is provided, lacking the requested specificity; 3. \textit{Partial} half-answers are provided; 4. \textit{Dodging} or ignoring the question altogether; and 5. \textit{Deflection} of the topic by shifting focus to a different point. The final category, named \textit{Clear Non-Reply}, includes \textit{Declining to Answer} in an explicit manner while acknowledging the question, \textit{Ignorance} when respondents admit they do not know the answer, and \textit{Clarification} when respondents ask for more details. 

An expert in linguistics, working professionally as an authorized translator, adapted the initial English taxonomy for Romanian, fixing category definitions and boundary cases. Appendix~\ref{app:taxonomy} reproduces the Romanian annotation guidelines. 

\subsection{Annotation Protocol}

We build a screening quiz of 15 QA pairs spanning ambiguous evasion categories, where each item is labeled by an expert to serve as a benchmark reference. The quiz is used as a screening step to ensure that the three annotators understand the taxonomy and can consistently handle borderline and difficult cases. All annotators are native Romanian women aged 22--24 with undergraduate training in psychology. 

Annotators complete a training phase that covers all nine categories of taxonomy (\autoref{sec:taxonomy}) using worked examples. Each instance requires a single annotation decision: assigning one of the nine evasion labels to the response. The three-way clarity label is then inferred by traversing the taxonomy upward. The annotation process spanned five weeks. During this period, we held three group review sessions with the expert to resolve disagreements and clarify boundary cases. Mean annotation time was 69 seconds per instance, totaling approximately 27 hours per annotator. All annotators were compensated for their work (see \autoref{sec:ethics}).

A single annotator labels each training instance, with the workload evenly distributed between the three. All three independently label the test split to support inter-annotator agreement measurement and evaluation using multiple annotations. For 2.4\% of test instances where all three annotators assigned different evasion labels, the expert assigned the final label.

\subsection{Worked Example}
\label{app:taxonomy:example}

\noindent
\textbf{Î:} \textit{``Domnule Președinte, pădurea este o chestiune de siguranță națională?''}\\
\textbf{R:} \textit{``Eu cred că pădurile sunt o chestiune de suflet național și de aceea sunt foarte încrezător că vom reuși să schimbăm lucrurile în bine.''}

\medskip
\noindent
\textit{English:}\\
\textbf{Q:} \textit{``Mr.\ President, are forests a matter of national security?''}\\
\textbf{A:} \textit{``I believe forests are a matter of national spirit, and for that reason I am very confident that we will succeed in changing things for the better.''}

\medskip
\noindent
The response receives clarity label \textit{Ambivalent} and evasion label \textit{Deflection}. The answer engages with the topic introduced in the question (forests as a national concern), but does not address whether forests constitute a national security issue. It reframes the discussion and shifts the attention away from the original premise without explicitly rejecting it.

\begin{figure}[ht]
\centering
\includegraphics[width=\linewidth]{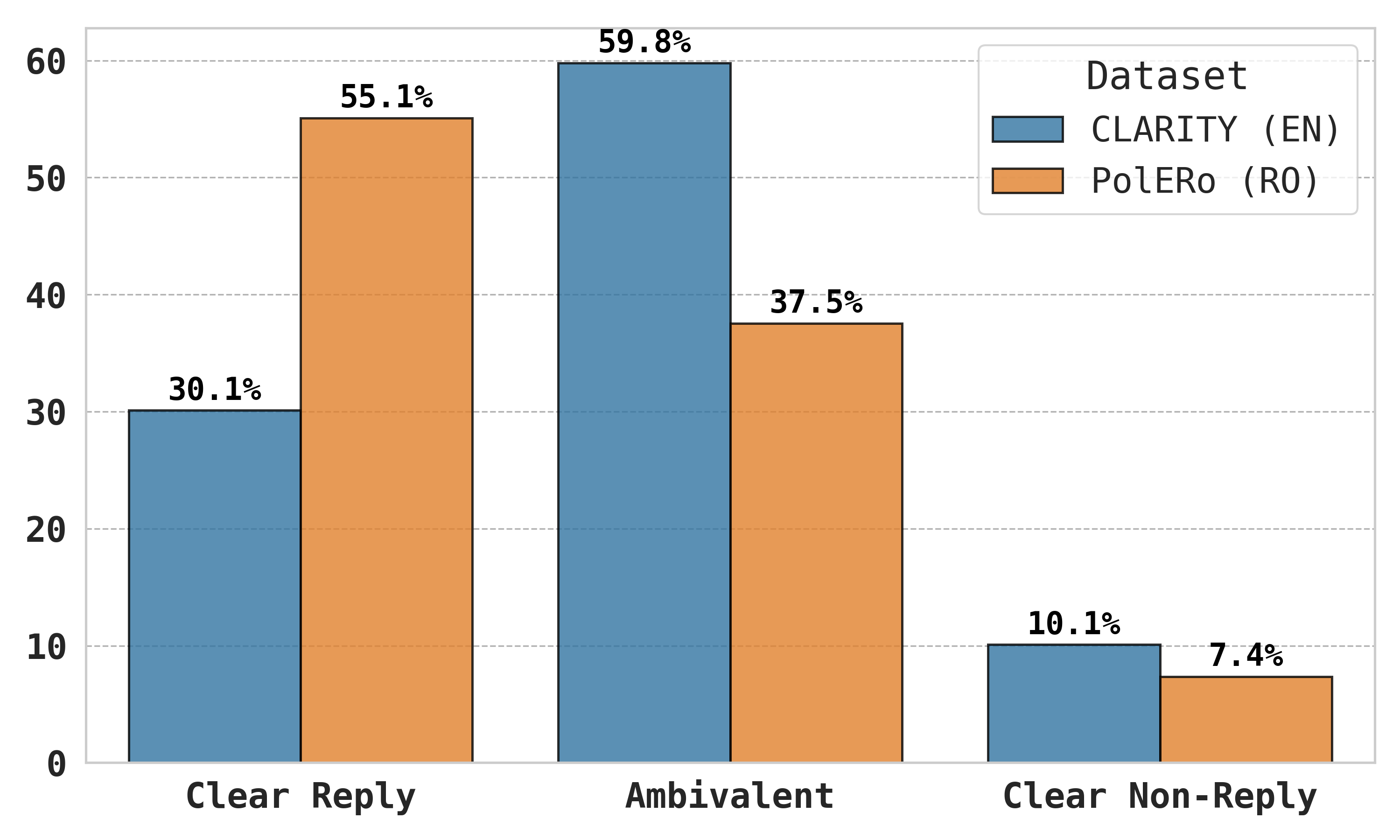}
\caption{Overall clarity distribution for PolERo compared to the English CLARITY dataset.}
\label{fig:main_clarity_dist}
\end{figure}

\subsection{Dataset Statistics}

Label distributions are similar across splits. In the training set, \textit{Clear Reply} accounts for 54.9\% of instances, \textit{Ambivalent} for 37.7\%, and \textit{Clear Non-Reply} for 7.4\%. The test set shows a comparable distribution (56.8\%, 36.1\%, 7.1\%).

Figure~\ref{fig:main_clarity_dist} compares the overall clarity label distribution of PolERo with the English CLARITY dataset \cite{thomas2024}. \textit{Clear Reply} is the majority class in the Romanian dataset (55.09\%), while the English corpus is primarily composed of \textit{Ambivalent} responses (59.80\%). Another point of departure between the two datasets is the way in which multi-part questions have been treated: in English, the interviews are decomposed into separate sub-questions using ChatGPT, while in Romanian, we discard such instances as we consider AI models insufficient for creating gold-standard data. An exploratory analysis is provided in the Appendix~\ref{app:eda}. The analysis reveals variation in response clarity across speakers, communicative settings, and topics. Romanian presidential discourse is dominated by \textit{Explicit} answers (55.09\%), although this proportion decreases in more formal contexts such as declarations and press conferences. Evasion strategies also vary across presidents and topics: earlier speakers rely more heavily on \textit{Dodging}, whereas post-2014 presidents exhibit more balanced distributions with higher rates of \textit{General}, \textit{Deflection}, and \textit{Declining to answer}. Finally, the length of the answer aligns with established theories of verbose evasion \citep{bavelas1988political, rasiah2010, bull2020can}, with indirect strategies such as \textit{General}, \textit{Deflection}, and \textit{Partial} responses tending to be substantially longer than direct responses.

\subsection{Annotation Quality}
To assess the performance of the three annotators, the expert independently labeled a benchmark set of 85 instances. Against this gold standard, the annotators achieved accuracies of 95.3\%, 95.3\%, and 94.1\% on the clarity task, and 89.4\%, 88.2\%, and 84.7\% on the evasion task (Appendix~\ref{app:annotation:benchmark}).

Inter-annotator agreement on the test set, measured using Fleiss's $\kappa$ \citep{fleiss1971}, is $0.843$ for clarity and $0.678$ for evasion, corresponding to near-perfect and substantial agreement on the scale of \citet{landis1977}. At the category level, agreement is highest for \textit{Clarification} ($\kappa=1.000$), \textit{Claims Ignorance} ($\kappa=0.881$), and \textit{Explicit} ($\kappa=0.858$), and lowest for \textit{Deflection} ($\kappa=0.364$) and \textit{General} ($\kappa=0.464$). A detailed analysis of the annotation is provided in Appendix~\ref{app:annotation}.

\section{Experimental Setup}
\label{sec:setup}

\subsection{Task Formulation}

We report macro-averaged F1 on the test split of each dataset as the primary metric across all experiments. We evaluate the models on two classification tasks derived from the two-level taxonomy (\autoref{sec:taxonomy}): (1)~\textit{clarity classification}, a three-class task over the coarse-grained labels; and (2)~\textit{evasion classification}, a nine-class task over the fine-grained labels. For the three-class clarity task, we also compute a macro-F1 score by deterministically (3) \textit{mapping} each predicted evasion label to its parent clarity class. Thus, we can directly compare models trained specifically for the three-class task against those whose coarse-grained labels are inferred through the fine-grained objective.

\begin{figure}[!t]
\centering
\resizebox{0.8\columnwidth}{!}{
\begin{tikzpicture}[
    node distance=0.8cm,
    block/.style={rectangle, draw, fill=blue!10, text width=2.5cm, text centered, rounded corners, minimum height=1cm},
    line/.style={draw, ->, >=latex, thick},
    chunk/.style={rectangle, draw, dashed, fill=gray!10, text width=1.5cm, minimum height=0.6cm, text centered}
]

\node (input) [text width=3.8cm, text centered] {Tokenized sequence \\ $T=\mathrm{tok}(Q \oplus A)$};

\node (chunk2) [chunk, below=0.5cm of input] {Chunk $1$};
\node (chunk1) [chunk, left=0.375cm of chunk2] {Chunk $0$};
\node (chunk3) [chunk, right=0.65cm of chunk2] {Chunk $M{-}1$};

\node (roberta) [rectangle, draw, fill=orange!20, minimum width=6.5cm, minimum height=0.8cm, below=0.6cm of chunk2]
{\textbf{Shared Encoder}};

\node (emb2) [below=0.6cm of roberta] {$h_1$};
\node (emb1) [left=1.5cm of emb2] {$h_0$};
\node (emb3) [right=1.5cm of emb2] {$h_{M-1}$};

\node (pool) [block, fill=green!10, below=0.6cm of emb2, text width=3.5cm]
{\textbf{Max-Pooling} \\ $v = \max_{k=0}^{M-1} h_k$};

\node (dropout) [block, fill=yellow!10, below=0.4cm of pool, text width=2.5cm] {Dropout};

\node (head_clarity) [block, fill=red!10, below left=0.3cm and -0.2cm of dropout]
{\textbf{Clarity} \\ (3 classes)};
\node (head_evasion) [block, fill=red!10, below right=0.3cm and -0.2cm of dropout]
{\textbf{Evasion} \\ (9 classes)};

\draw[line] (input) -- (chunk2);

\draw[dashed] (chunk1.south) -- (chunk1.south |- roberta.north);
\draw[dashed] (chunk2.south) -- (chunk2.south |- roberta.north);
\draw[dashed] (chunk3.south) -- (chunk3.south |- roberta.north);

\draw[line] (roberta.south -| emb1) -- (emb1);
\draw[line] (roberta.south -| emb2) -- (emb2);each chunk is encoded by a shared encoder, chunk repre-
\draw[line] (roberta.south -| emb3) -- (emb3);

\draw[line] (emb1) -- (pool);
\draw[line] (emb2) -- (pool);
\draw[line] (emb3) -- (pool);

\draw[line] (pool) -- (dropout);
\draw[line] (dropout) -| (head_clarity);
\draw[line] (dropout) -| (head_evasion);

\end{tikzpicture}
}
\caption{Multi-head architecture. The tokenized concatenated input ($Q\oplus A$) is split into overlapping chunks, each chunk is encoded by a shared encoder, chunk representations are aggregated via element-wise max-pooling, and two task-specific heads predict clarity (3 classes) and evasion (9 classes).}
\label{fig:multihead_arch}
\end{figure}
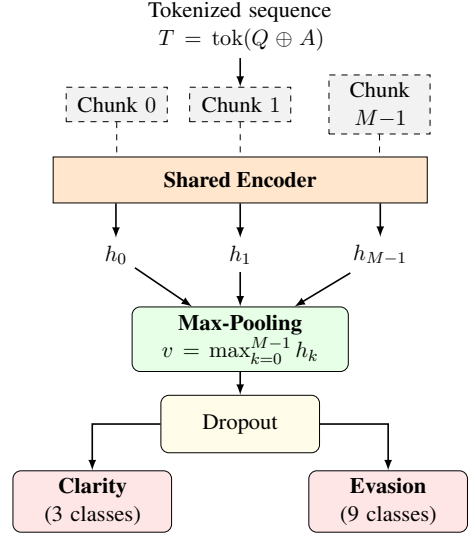

\subsection{Models}

We establish baselines across four model families on both the English CLARITY dataset and PolERo, training and evaluating each language independently. Input formats, hyperparameters, and prompt templates are reported in Appendix~\ref{app:experiments:methodology}. A TF-IDF logistic regression classifier serves as a non-neural lower bound. The model comparison covers single-head (SH) encoder fine-tuning, a custom multi-head (MH) encoder rendered in \autoref{fig:multihead_arch}, and  LLMs under zero-shot and few-shot prompting without fine-tuning: GPT-5.4, Llama-3.3-70B-Instruct, Qwen3.6-35B-A3B, Gemma-4-31B-it, DeepSeek-V4-Pro, and gpt-oss-120b. 

Standard encoder configurations are finetuned independently for each task. In English, we evaluate RoBERTa-large \citep{liu2019roberta}, 
ModernBERT-large \citep{warner-etal-2025-smarter}, ELECTRA-large \citep{clark2020electrapretrainingtextencoders}, BERT-large-cased \citep{devlin-etal-2019-bert}, and XLM-RoBERTa-large \citep{conneau-etal-2020-unsupervised}. The same five architectures are fine-tuned on PolERo, with the addition of \textbf{Ro}BERT-large \citep{masala-etal-2020-robert},\footnote{We emphasize the first two letters of Romanian BERT large: \textbf{Ro}BERT-large to increase readability and distinguish it more easily from the RoBERTa-large model, both used extensively in this work.} and Romanian BERT \citep{dumitrescu-etal-2020-birth}, both pretrained on Romanian corpora, and multilingual mBERT-base \citep{devlin-etal-2019-bert}.

The Multi-head (MH) encoder architecture contains a shared encoder and two linear heads that predict clarity and evasion jointly (Figure~\ref{fig:multihead_arch}). To process long question-answer pairs without truncation, sequences are segmented into overlapping chunks and encoded independently. The chunk representations are aggregated via element-wise max-pooling into a single response vector, which is passed to a 3-class clarity head and a 9-class evasion head. The training objective is the unweighted sum of the two cross-entropy losses. Models are trained with 7-fold stratified cross-validation, and fold predictions are ensembled by probability averaging at inference time. Matching the single-head configurations, we train RoBERTa-large, XLM-RoBERTa-large (added to support the cross-lingual experiments below), and \textbf{Ro}BERT-large, a Romanian-only encoder model. Full architectural details and mathematical formulations together with component ablations isolating the contribution of chunked pooling, joint supervision, and ensembling are reported in Appendix~\ref{app:multihead}.

\section{Results and Discussion}
\label{sec:results}

The results in this work are reported on the English dataset using the \textit{development} set from the SemEval 2026 CLARITY Shared Task, since the official test of the task was not publicly available at the time of writing.  The multi-head RoBERTa-large achieves one of the strongest scores (0.8) on the SemEval 2026 CLARITY task \cite{sgunibuc, thomas2026clarity} among encoder-only models, especially given the size of under 400 million parameters.\footnote{According to the official report by \citet{thomas2026clarity}, all top-scoring systems in the range of $0.82$--$0.89$ relied on proprietary LLMs.} However, on the development set, the same model obtains a smaller $.713$ score for the 3-way clarity task prediction.

Table~\ref{tab:main_summary} reports the macro-F1 scores for the strongest models from each approach. A complete set of results for all models, cross-lingual setups, and prompting strategies is provided in Appendix~\ref{app:experiments:comparison}.

\paragraph{Traditional classifiers.}
TF-IDF + logistic regression reaches a macro-F1 $.422$/$0.453$ on English/Romanian 3-class clarity detection, confirming that surface-level lexical features alone are insufficient. Additional per-class error patterns are reported in Appendix~\ref{app:experiments}.

\paragraph{Monolingual encoder fine-tuning.}
RoBERTa-large is the strongest single-head English encoder model for the 3-class clarity (C) identification task ($.580$) and the 9-class fine-grained evasion (E) detection task ($.481$). The mapped (M) fine-grained task significantly outperforms ($.656$) the simple clarity identification task, showing that it is better to train on a more fine-grained taxonomy and merge than to train directly on fewer un-balanced classes. 

Unsurprisingly, for Romanian, this model performs significantly worse than encoder models pre-trained on Romanian, such as \textbf{Ro}BERT-large \cite{masala-etal-2020-robert}. The latter is the strongest single-head model on PolERo ($.504$ evasion, $.701$ mapped). More detailed encoder comparisons are available in Appendix~\ref{app:experiments:comparison}.

\begin{table}[ht]
\centering
\resizebox{\linewidth}{!}{%
\begin{tabular}{l ccc ccc}
\toprule
& \multicolumn{3}{c}{\textbf{EN (CLARITY)}} & \multicolumn{3}{c}{\textbf{RO (PolERo)}} \\
\cmidrule(lr){2-4}\cmidrule(lr){5-7}
\textbf{Model} & C & E & M & C & E & M \\
\midrule
TF-IDF + LR & .422 & .301 & .441 & .453 & .312 & .472 \\
\midrule
\multicolumn{7}{l}{\textit{Best Encoders (Fine-Tuned)}} \\
RoBERTa (SH)                & .580 & .481 & .656 & .646 & .428 & .613 \\
RoBERTa (MH)  & \textbf{.713} & .568 & \textbf{.731 }& .696 & .501 & .652 \\
XLM-R. (SH)           & .579 & .419 & .564 & .680 & .365 & .686 \\
XLM-R. (MH)          & .665 & .475 & .644 & .694 & .508 & .706 \\
\textbf{Ro}BERT (SH)            & --   & --   & --   & .688 & .504 & .701 \\
\textbf{Ro}BERT (MH)   & --   & --   & --   & \textbf{.729} & .545 & \textbf{.763} \\
\midrule
\multicolumn{7}{l}{\textit{Best LLMs (Few-Shot Prompting)}} \\
Llama-3.3-70B (few-shot)         & .535 & .382 & .564 & .555 & .312 & .411 \\
GPT-OSS-120B (few-shot)          & .542 & .554 & .729 & .553 & .580 & .729 \\
DeepSeek-V4-Pro (few-shot)       & .601 & .545 & .670 & .627 & .594 & .751 \\
Gemma-4-31B-it & \textbf{.721} & .583 & .662 & .672 & .619 & .738 \\
Qwen3.6-35B-A3B   & .669 & .544 & .747 & .746 & \textbf{.682} & .760 \\
GPT-5.4        & .703 & \textbf{.626} & \textbf{.761} & \textbf{.763} & .666 & \textbf{.763} \\
\bottomrule
\end{tabular}%
}
\caption{Summary of test-set macro-F1 for the highest-performing models across all tested approaches. \textbf{C}: Clarity (direct 3-class prediction). \textbf{E}: Evasion (direct 9-class prediction). \textbf{M}: Mapped (3-class clarity prediction obtained by mapping the predicted 9-class evasion label upward through the taxonomy). \textbf{SH}: Single-head, \textbf{MH}: Multi-head. Best scores are highlighted in \textbf{bold}. LLMs obtain the strongest results for the fine-grained Evasion prediction, while smaller-sized multi-head encoders remain competitive for predicting response clarity.}
\label{tab:main_summary}
\end{table}

\paragraph{Multi-head fine-tuning.}
The custom multi-head model architecture improves performance across all metrics compared to the corresponding single-head model on both datasets. Multi-head RoBERTa-large is the strongest English encoder configuration, outperforming its single-head counterpart by $13$ macro-F1 points on direct clarity and 9 on evasion, and surpassing GPT-5.4 on direct clarity prediction. 
The multi-head \textbf{Ro}BERT-large improves over its single-head Romanian-specific counterpart by $4$ points on clarity, $4$ on evasion, and $6$ on mapped clarity. These gains are consistent with the dual clarity+evasion tasks acting as a regularizer: gradients from the coarser clarity supervision help stabilize predictions for rare evasion categories, and vice-versa.

Multi-head XLM-RoBERTa-large obtains smaller gains in both languages than the language-specialized backbones, indicating that multilingual pretraining sacrifices some within-language performance for broader cross-language coverage. As expected, models pretrained specifically for a given language consistently outperform multilingual models on that language: \textbf{Ro}BERT-large achieves higher scores than XLM-RoBERTa-large on all Romanian evaluation metrics, while RoBERTa-large similarly outperforms XLM-RoBERTa-large on the English benchmarks.

\begin{table}[htb]
\centering
\small
\renewcommand{\arraystretch}{1.15}
\begin{tabular*}{\columnwidth}{@{\extracolsep{\fill}}ll rr@{}}
\toprule
\textbf{Step} & \textbf{Component} & \textbf{Clarity} & \textbf{Evasion} \\
\midrule
\multicolumn{4}{@{}l}{\textit{English (CLARITY), RoBERTa-large}} \\
S0 & Single-head       & $.597_{\pm.006}$ & $.474_{\pm.024}$ \\
S1 & Multi-head        & $.638_{\pm.007}$ & $.478_{\pm.013}$ \\
S2 & Sliding-window    & $.648_{\pm.024}$ & $.481_{\pm.040}$ \\
S3 & Cross-validation  & $.662_{\pm.017}$ & $.513_{\pm.016}$ \\
S4 & Ensembling        & $.694_{\pm.009}$ & $.523_{\pm.013}$ \\
\cmidrule(l{2pt}r{2pt}){1-4}
\multicolumn{2}{@{}l}{\textit{Gain over S0}} & $\mathbf{+.097}$ & $\mathbf{+.049}$ \\
\midrule
\multicolumn{4}{@{}l}{\textit{Romanian (PolERo), \textbf{Ro}BERT-large}} \\
S0 & Single-head       & $.678_{\pm.022}$ & $.511_{\pm.015}$ \\
S1 & Multi-head        & $.701_{\pm.022}$ & $.531_{\pm.030}$ \\
S2 & Sliding-window    & $.705_{\pm.017}$ & $.515_{\pm.019}$ \\
S3 & Cross-validation  & $.707_{\pm.015}$ & $.516_{\pm.046}$ \\
S4 & Ensembling        & $.721_{\pm.011}$ & $.537_{\pm.021}$ \\
\cmidrule(l{2pt}r{2pt}){1-4}
\multicolumn{2}{@{}l}{\textit{Gain over S0}} & $\mathbf{+.043}$ & $\mathbf{+.026}$ \\
\bottomrule
\end{tabular*}
\caption{Cumulative ablation of the multi-head architecture.}
\label{tab:ablation_ladder}
\end{table}


Table~\ref{tab:ablation_ladder} adds each component cumulatively to a single-head baseline, repeating the full encoder pipeline for both languages, with RoBERTa-large as the English backbone and \textbf{Ro}BERT-large as the Romanian one. Each configuration is run over three random seeds (7, 42, 123); we report mean and standard deviation of test-set macro-F1 together with the increment over the preceding step. Relative to the single-head baseline, the complete architecture gains $+0.097$ macro-F1 on English clarity, $+0.049$ on English evasion, $+0.043$ on Romanian clarity and $+0.026$ on Romanian evasion. The joint multi-head objective provides the largest single gain on clarity ($+0.041$ English, $+0.023$ Romanian), while cross-validation and ensembling account for most of the improvement on the evasion task ($+0.032$, $+0.010$ on English).

\paragraph{Predicting mapped vs.\ direct clarity.}
The two-step strategy of predicting the nine fine-grained evasion categories and then projecting upward through the taxonomy outperforms the direct clarity head for 
RoBERTa-large on English, for \textbf{Ro}BERT-large on Romanian, and for nearly every LLM configuration on both datasets. This is consistent with the results reported at the Shared Task \cite{thomas2026clarity}, where \textit{the winning reverse derivation strategy proved to be the single most effective approach across the entire competition}.

Four encoders do not benefit from this approach: mBERT-base, BERT-large-cased, ELECTRA-large, and ModernBERT-large. In each case, the evasion head assigns most predictions to \textit{Explicit} and \textit{General}, causing the projection step to overpredict \textit{Clear Reply} at the expense of \textit{Ambivalent} and \textit{Clear Non-Reply}. ELECTRA-large on the Romanian split is the clearest example: the model predicts \textit{Explicit} for every instance, yielding a macro-F1 of $.084$, and the mapped clarity score 
drops to $.24$ because \textit{Ambivalent} and \textit{Clear Non-Reply} are never predicted. ModernBERT-large shows the same pattern, though less severely, with two categories still receiving zero recall on Romanian (\textit{Claims ignorance} and \textit{Partial/half-answer}). The two-step strategy therefore depends on the backbone being able to assign a meaningful probability mass to rare categories. Otherwise, the projection step amplifies the model's bias toward the majority class rather than correcting it.

\paragraph{LLM prompting.}
LLMs achieve some of the best results across all metrics on both datasets, though their ranking varies by task, and there are many cases when models perform significantly weaker than encoders (\autoref{tab:main_summary}). Here we report only the strongest candidates; for a full set of results, see Appendix \ref{app:llmpromptanalys} and \autoref{tab:full_results}. Gemma-4-31B-it obtains the best few-shot scores on direct clarity prediction for English, Qwen3.6-35B-A3B achieves the best score on few-shot evasion for Romanian, and GPT-5.4 attains strong mapped clarity scores on both, although it is on-par with the multi-head \textbf{Ro}BERT model. 

DeepSeek-V4-Pro and gpt-oss-120B fall between these models. Few-shot prompting improves the mapped metric for most models because the examples help clarify the boundaries between similar categories, especially \textit{Dodging}-\textit{Deflection} and \textit{Implicit}-\textit{General}. Under zero-shot prompting, these are the category pairs that models commonly confuse, and even one example per category substantially improves alignment. Llama-3.3-70B is the exception. Its mapped clarity score on English drops from $.624$ to $.564$ under few-shot prompting. An analysis of the outputs reveals that the model becomes heavily biased toward \textit{Deflection}, predicting it for 79.5\% of all English test instances, regardless of the gold label. The same holds for Romanian, where \textit{Deflection} accounts for 44\% of all predictions. In our experiments, the 70B Llama model obtained relatively weak results, which was unexpected given that it is one of the strongest baselines at the SemEval Task 6 CLARITY \cite{thomas2026clarity}.

\paragraph{Data contamination.}
The performance differences between LLMs and fine-tuned encoders likely stem from multiple factors. Firstly, LLMs benefit from broad parametric world knowledge stored in large billion-parameter-sized networks, which help to resolve entity-dependent evasion. Secondly, all evaluated LLMs (except Llama-3.3-70B) are prompted with reasoning enabled, allowing for potentially better deliberation over pragmatic intent. Notably, Llama-3.3-70B, which lacks reasoning, is also the weakest LLM in both languages. Thirdly, we cannot rule out that these models might have been exposed to the data (which is publicly available) during their pre-training phase. To assess whether the advantage of LLMs is influenced by data contamination, we explore the CoDeC framework,  Contamination Detection via Completion \cite{zawalski2026detectingdatacontaminationllms}, on three open-source LLMs: Qwen3.6-35B-A3B, DeepSeek-V4-Pro, and Gemma-4-31B-it. CoDeC scores are extracted solely from the text, yielding contamination percentages on the English test set of $37\%$, $38\%$, and $45\%$; while on the Romanian test set, the corresponding scores are $43\%$, $41\%$, and $39\%$. All six values fall below the $80\%$ threshold that would indicate a clear case of data contamination, according to the \citet{zawalski2026detectingdatacontaminationllms}. Even if the models could have been exposed to the same texts during their pre-training (this includes the encoders), the associated labels were not openly available; therefore, we could not find sufficient evidence for contamination.

\section{Cross-Lingual Transfer}
\label{sec:experiments:crosslingual}

We assess cross-lingual generalization using \textit{the multi-head (MH) architecture} and XLM-RoBERTa-large and RoBERTa-large, although the latter is reserved for Appendix, \autoref{tab:full_results} since it is not designed to be a multilingual encoder.

Each model is trained in multiple cross-lingual configurations:

\begin{itemize}[noitemsep,topsep=2pt]
    \item \textbf{EN $\rightarrow$ RO}: trained on English, evaluated on Romanian.
    \item \textbf{RO $\rightarrow$ EN}: trained on Romanian, evaluated on English.
    \item \textbf{EN + RO $\rightarrow$ RO} and \textbf{$\rightarrow$ EN}: joint training on the concatenation of both datasets, evaluated only on Romanian or English.
\end{itemize}

The first two configurations explore the ability to generalize across the two languages and political spheres (U.S. vs. Romania), while the remaining configurations explore the possibility of increasing the predictive power using multilingual data augmentation. Results are available in \autoref{tab:mtsummary}.

Training on English data and predicting on Romanian yields a significant 6 to 7 point drop ($.694 \rightarrow .635$) using both XLM- and RoBERTa-large models. 
Training in the opposite direction, in Romanian $\rightarrow$ English, results in a worse performance degradation of 11 to 14 points. 
In both scenarios, the performance degradation makes the MH models similar to other single-head models such as mBERT-base, BERT-large-cased, ELECTRA-large, or ModernBERT (see \autoref{tab:full_results} containing complete experimental results in the Appendix).

The model trained on Romanian strongly over-predicts \textit{Clear Reply}, misclassifying 54\% of English \textit{Ambivalent} answers as \textit{Clear Reply}, and similarly maps most English \textit{Implicit} and \textit{Deflection} responses to \textit{Explicit}. These results indicate that cross-lingual predictions do not lead to catastrophic losses and that the task may benefit from additional multilingual data.


\paragraph{Multilingual augmentation.}
As shown in \autoref{tab:mtsummary}, combining the two datasets and predicting on each test set leads to improvements over individual training of the same model.
One of the best configurations for Romanian is joint EN+RO training with XLM-RoBERTa-large, which achieves the highest Romanian evasion score among fine-tuned models ($0.570$). This may be the case because joint training regularizes rare categories: Romanian examples such as \textit{Clarification} or \textit{Claims ignorance} are reinforced by similar English instances, reducing the data scarcity that causes failures. For example, \textit{Declining to answer} reaches perfect recall on the Romanian test split under joint training. Furthermore, joint training encourages the shared encoder to model evasion in a cross-lingual way rather than relying on language-specific lexical cues. The English dataset contains more \textit{Ambivalent} evasive answers, where evasion is expressed through topic shifts and intent rather than specific words. When trained jointly, the model may transfer this ability to Romanian, improving its ability to detect evasion without clear lexical signals. As a result, Romanian evasion scores increase by $0.062$ over the in-language XLM-RoBERTa-large baseline, whereas the English scores show an insignificant improvement (+$0.008$).

RoBERTa-large does not benefit in the same way (\autoref{tab:full_results}). Its tokenizer and pretraining are optimized for English, so adding Romanian data introduces more noise than useful signal, causing English evasion performance to drop from $0.568$ to $0.503$. Overall, the English performance drops consistently across all criteria when doing joint training.

\subsection{Machine Translation}
As a secondary analysis, we extend the cross-lingual setup with machine-translated training and test data in order to reduce the biases of the multilingual encoders. All translations are generated with NLLB-200-distilled-600M \citep{nllbteam2022languageleftbehindscaling}. Appendix~\ref{app:mt_quality} reports a semi-automatic analysis of the quality estimation of the translations using XCOMET-XL \citep{guerreiro-etal-2024-xcomet}. The model indicates higher MT quality for RO$\rightarrow$EN than EN$\rightarrow$RO, although manual inspection shows that many EN$\rightarrow$RO translations flagged as errors are acceptable, with pragmatic categories such as \textit{Deflection} remaining the most error-prone in both translation directions.

\begin{table}[ht]
\centering
\resizebox{\linewidth}{!}{%
\begin{tabular}{l ccc ccc}
\toprule
& \multicolumn{3}{c}{\textbf{EN (CLARITY)}} & \multicolumn{3}{c}{\textbf{RO (PolERo)}} \\
\cmidrule(lr){2-4}\cmidrule(lr){5-7}
\textbf{Model} & C & E & M & C & E & M \\
\midrule
\multicolumn{7}{l}{\textit{Best encoders fine-tuned on the training set}} \\
\textbf{Ro}BERT-large (MH)            & --   & --   & --   & .729 & .545 & \textbf{.763} \\
RoBERTa (MH)  & .713 & .568 & .731  & .-- & .-- & .-- \\
\midrule
\multicolumn{7}{l}{\textit{Cross-lingual XLM-R multi-head models:}} \\
\midrule
Individual Training          & .665 & .475 & .644 & .694 & .508 & .706 \\
Joint Training (EN+RO)           & .679 & .483 & .642 & \textbf{.723} & \textbf{.570} & .717 \\
\midrule
\multicolumn{7}{l}{\textit{Low-data scenario; only test set available}} \\
EN $\rightarrow$ RO                                  & --   & --   & --   & .635 & .399 & .655 \\
EN2RO $\rightarrow$ RO                               & --   & --   & --   & .665 & .412 & .706 \\
RO $\rightarrow$ EN                                  & .551 & .398 & .527 & --   & --   & --   \\
RO2EN $\rightarrow$ EN                               & .477 & .358 & .439 & --   & --   & --   \\
\midrule
\multicolumn{7}{l}{\textit{MT augmentation}} \\
RO + EN2RO $\rightarrow$ RO                          & --   & --   & --   & .718 & .489 & .684 \\
EN + RO2EN $\rightarrow$ EN                          & .694 & .477 & .602 & --   & --   & --   \\
\midrule
\multicolumn{7}{l}{\textit{Best LLMs (Few-Shot Prompting)}} \\
Gemma-4-31B-it & \textbf{.721} & .583 & .662 & .672 & .619 & .738 \\
Qwen3.6-35B-A3B   & .669 & .544 & .747 & .746 & \textbf{.682} & .760 \\
GPT-5.4        & .703 & \textbf{.626} & \textbf{.761} & \textbf{.763} & .666 & \textbf{.763} \\
\bottomrule
\end{tabular}%
}
\caption{The first two and the last three rows contain the best models for each language and task. The arrow $\rightarrow$ shows the training and testing direction of the XLM-R, multi-head models. The labels EN2RO and RO2EN indicate machine translated training sets. In low-data scenarios it is best to use few shot prompting and LLMs. When data is available, joint multi-lingual training achieves better results than monolingual training using machine translated data.}
\label{tab:mtsummary}
\end{table}

Translating the English data into Romanian partially recovers cross-lingual clarity performance but provides limited gains for evasion detection, according to the results presented in \autoref{tab:mtsummary}. Categories such as \textit{Deflection} and \textit{Dodging} suffer the largest degradation under MT augmentation and also receive the lowest translation quality scores, whereas more formulaic categories such as \textit{Clarification} and \textit{Declining to answer} translate reliably. A transfer asymmetry between English and Romanian is visible --- machine translating texts from Romanian to English degrades the results more than the other way around. We believe that differences in class distributions and different discourse structures, shaped by historical practices, geopolitical contexts, and topics of interest, make the Romanian data less useful for detecting evasion in U.S. English. MT-based training set augmentation provides inconclusive results compared to individual or joint multilingual training.

\section{Remarks}
Based on our Romanian case study, we propose the following takeaways to guide future assessments of political evasion in new languages:
\begin{enumerate}[noitemsep,topsep=3pt]
    \item In low-resource settings, when creating a dataset for a new language such as Romanian, it is advisable to first benchmark the dataset using few-shot prompting with available LLMs, preferably open-weight models.
    \item If LLMs perform poorly for a particular language, translating existing datasets with NLLB-200 or similar tools and training cross-lingual encoders may be a viable alternative, although factors related to data representativeness \cite{dogruoz-etal-2023-representativeness}, geopolitical factors, and local histories must be carefully taken into account.
    \item If a sufficiently large training set can be constructed for the target language (i.e., at least 3,000 samples) and existing monolingual pre-trained encoders are available, then a strong strategy is to use the multi-head approach described in Section~\ref{sec:setup}; otherwise, if monolingual encoders are not available, then joint multilingual training may be more beneficial than augmenting the data with machine-translated examples.
    \item The strong performance of LLMs must be weighed against cost and efficiency; our proposed multi-head encoder proved competitive with LLMs on the clarity classification task while using fewer than 500M parameters; at the same time, the encoders have limitations in terms of fine-grained evasion detection.
    \item The distribution of ambiguous responses can have a significant effect on the generalizability of cross-lingual models. 
\end{enumerate}

\section{Conclusions}
In this work, we introduce \textit{PolERo}, a novel Romanian dataset of $3{,}574$ question-answer pairs from presidential transcripts annotated using a two-level taxonomy of evasion. We compare multiple models on an existing English dataset and on our proposed Romanian dataset under matched conditions, covering traditional classifiers, standard encoder fine-tuning, a custom multi-head architecture for joint clarity and evasion prediction, and large language models under zero-shot and few-shot prompting.

Our results only partially corroborate previous findings \cite{thomas2026clarity} that prompting modern LLMs strongly outperforms fine-tuned encoders. We show that highly competitive results can be obtained on a language such as Romanian using encoder models only. We propose a sliding-window and max-pooling approach with two classification heads that significantly increases the predictive power of encoder models, making them competitive with LLMs for the three-class clarity classification task while using only a fraction of the memory footprint. However, for the nine-class fine-grained evasion task, larger models achieve better predictive performance.

Finally, we investigate cross-lingual transfer from the perspective of multilingual training versus monolingual training with machine-translated texts. Our results reveal an asymmetric relation: English-trained models generalize better to Romanian than the other way around. Joint multilingual training yields good cross-lingual performance for Romanian, while ambivalent categories remain the main challenge across all model families, as they depend on pragmatic intent rather than surface form.

\section{Limitations}
PolERo covers a single non-English language. Whether the taxonomy, annotation difficulty, and model behavior observed here generalize to other languages and political contexts remains an open question.


The annotation protocol assigns a single annotator per training instance, with triple annotation reserved for the test split. Evasion classification requires interpreting communicative intent, which can vary across annotators. Our annotators have undergraduate training in psychology rather than political science or linguistics, which may affect interpretation of borderline cases. Full triple annotation with domain experts would likely reduce this uncertainty, but at substantially higher cost.

\section{Ethical Considerations}
\label{sec:ethics}
Source data consists of official transcripts published on the public archives of the Romanian presidency. The speakers are public officials whose statements were issued in their official capacity; no private communications, personal data, or non-public records are included.

The three annotators were each paid the equivalent of 80~EUR for approximately 27 hours of labeling work distributed over five weeks. The linguistic expert contributed voluntarily. Before participation, annotators were informed of the task purpose, the political nature of the content, and the option to withdraw at any point without penalty.

Clarity and evasion classification of political speech can be used to selectively frame or attack individual speakers. We release PolERo for research on response clarity in political discourse and discourage its deployment as a standalone instrument of public accountability. Model predictions are not ground truth: as the inter-annotator agreement statistics show, even trained human annotators disagree on a fraction of instances. Automated labels should not be treated as definitive characterizations of any individual speaker's behavior.

Finally, the computational cost associated with the experimental pipeline presented in this work, covering fine-tuning and inference experiments, was conducted on a single NVIDIA H100 80GB GPU, with a total estimated budget of approximately 180 GPU hours. AI was used to check grammar and to generate boilerplate code.

\section*{Acknowledgments}
We would like to thank the reviewers for their helpful comments, and in particular, reviewer NvEV for their creative suggestions on how this work can be used as a political tool.

This research is partially supported by the project “Romanian Hub for Artificial Intelligence - HRIA”, Smart Growth, Digitization and Financial Instruments Program, 2021-2027, MySMIS no. 351416, and by InstRead: Research Instruments for Text Complexity, Simplification, and Readability Assessment CNCS - UEFISCDI project number PN-IV-P2-2.1-TE-2023-2007.


\bibliography{main}

\appendix

\section{Taxonomy and Annotation Guidelines}
\label{app:taxonomy}

\subsection{Romanian Annotation Guidelines}
\label{app:taxonomy:romanian}

\paragraph{Label names.} We retained the nine original English label names (\textit{Explicit}, \textit{Implicit}, \textit{General}, \textit{Partial/half-answer}, \textit{Dodging}, \textit{Deflection}, \textit{Declining to answer}, \textit{Claims ignorance}, \textit{Clarification}) to ensure our dataset remains directly comparable with the English corpus.

\paragraph{Descriptions and examples.} We translated all category descriptions into Romanian and replaced the English examples of \citet{thomas2024} with Romanian examples drawn from political discourse data in our dataset. Each example pairs an answer (R) with a question (Î) and a short rationale identifying the features that trigger the label.

\noindent A Romanian linguistic expert assisted with example selection to ensure that each category is illustrated by a clear and representative instance. Table~\ref{tab:taxonomy_full} reproduces the guidelines, with English descriptions and bilingual examples (Romanian as supplied to annotators; English translation in italics for readability).

\subsection{Summaries and Quality Control}
\label{app:taxonomy:quality_control}

Annotators were required to classify the evasion strategy based on the original text. In addition, the annotators received GPT-5.4-generated summaries and were asked to verify the validity of the summaries. To ensure that the annotators were reading the source text, we inserted 60 intentionally incorrect summaries. The annotators correctly flagged $58$ of these ($96.67\%$), producing only two false negatives. None of the other summaries were flagged as inconsistent.

\section{Annotation Protocol}
\label{app:annotation}

\subsection{Labeling Platform}
\label{app:annotation:platform}

\begin{figure}[ht]
\centering
\includegraphics[width=\columnwidth]{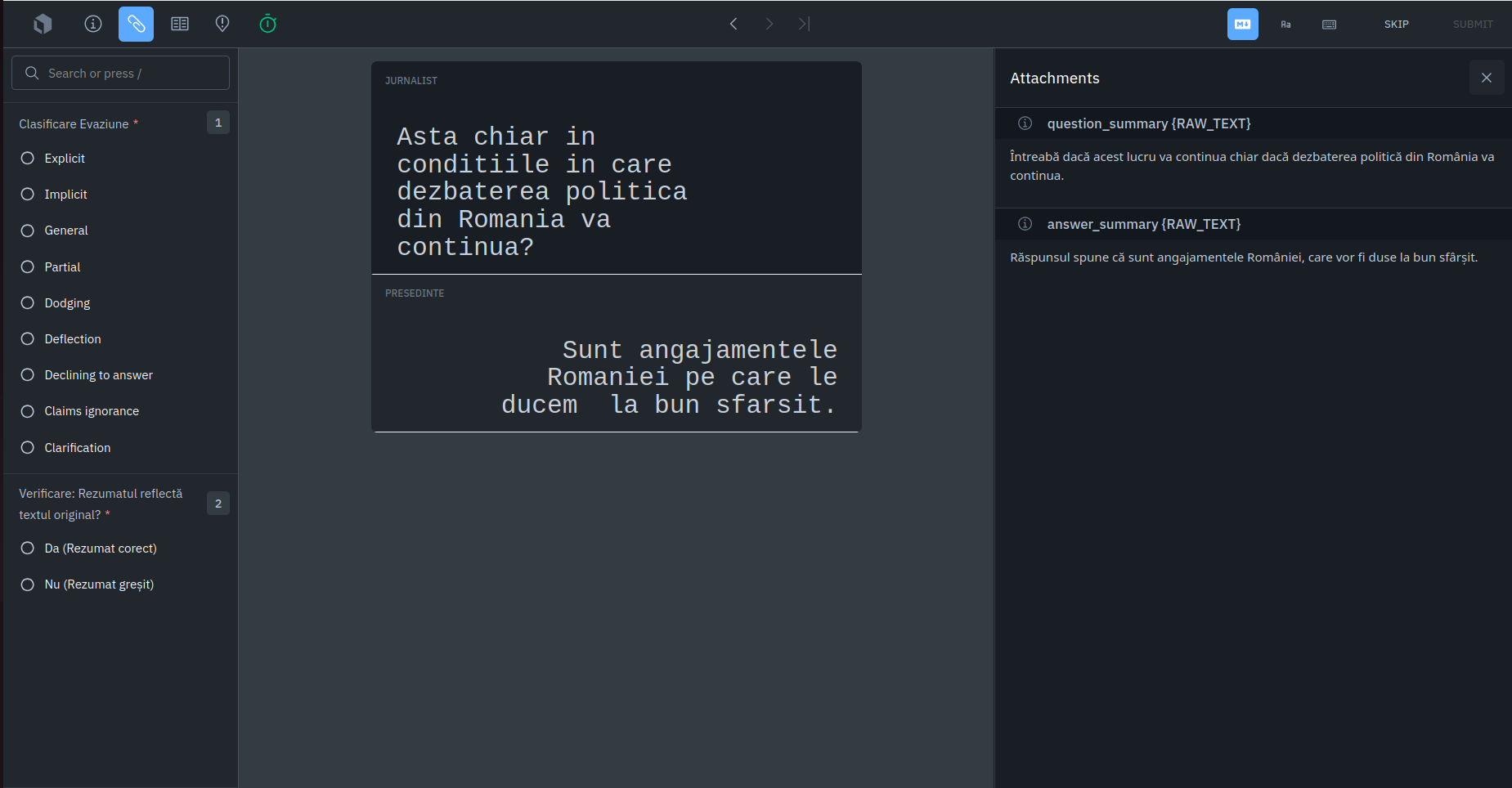}
\caption{Labelbox annotation interface. Left panel: evasion classification and summary validity check. Center: original question and answer. Right panel: GPT-5.4-generated question and answer summaries as attachments.}
\label{fig:labelbox_interface}
\end{figure}

We used Labelbox\footnote{\url{https://labelbox.com/}} to manage the annotation workflow. Each item displayed the original interview question and answer verbatim, with the GPT-5.4-generated question and answer summaries attached as side panels for reference (Figure~\ref{fig:labelbox_interface}). The annotator completed two tasks per item: (i)~assign one of the nine evasion labels from the Level-2 taxonomy, and (ii)~verify whether the answer summary correctly reflects the original. 

The platform provided the complete annotation guidelines, which included the taxonomy table (Table~\ref{tab:taxonomy_full}) and an extended set of examples for border-case examples covering categories with closely related semantics (e.g.\ \textit{Dodging} vs.\ \textit{Deflection}, \textit{General} vs.\ \textit{Partial/half-answer}). Annotators could consult the guidelines at any point during a labeling session.

Annotators could additionally tag any instance with one of five issue categories:

\begin{itemize}\itemsep0pt
\item \textit{Parsing Error} --- the QA pair was malformed during extraction.
\item \textit{Not a Question} --- the journalist's statement does not solicit any information.
\item \textit{Multiple Questions} --- a single journalist's question contains several sub-questions.
\item \textit{Insufficient Context} --- the response cannot be accurately classified without additional background information.
\item \textit{Borderline / Hard to Decide} --- the response is difficult to classify and may fall between two evasion categories.
\end{itemize}

\noindent Annotators could also create custom issue categories to report any additional problems not covered by the predefined taxonomy.

\noindent Items tagged with the first three issues were filtered from the released corpus. Items tagged \textit{Insufficient Context} or \textit{Borderline / Hard to Decide} were escalated to the expert, who either adjudicated a final label or removed the instance.

\subsection{Inter-Annotator Agreement}
\label{app:annotation:iaa}

Raw annotator agreement on the test set is high, especially for clarity. All three annotators agree on 87.2\% of clarity items, while the remaining 12.8\% receive a two-thirds majority, leaving no items without consensus. Agreement on evasion is slightly lower but still substantial: 69.6\% of items receive full three-way agreement, 28.0\% receive a two-thirds majority, and only 2.4\% result in complete disagreement between all three annotators.

\begin{figure}[ht]
\centering
\includegraphics[width=0.95\columnwidth]{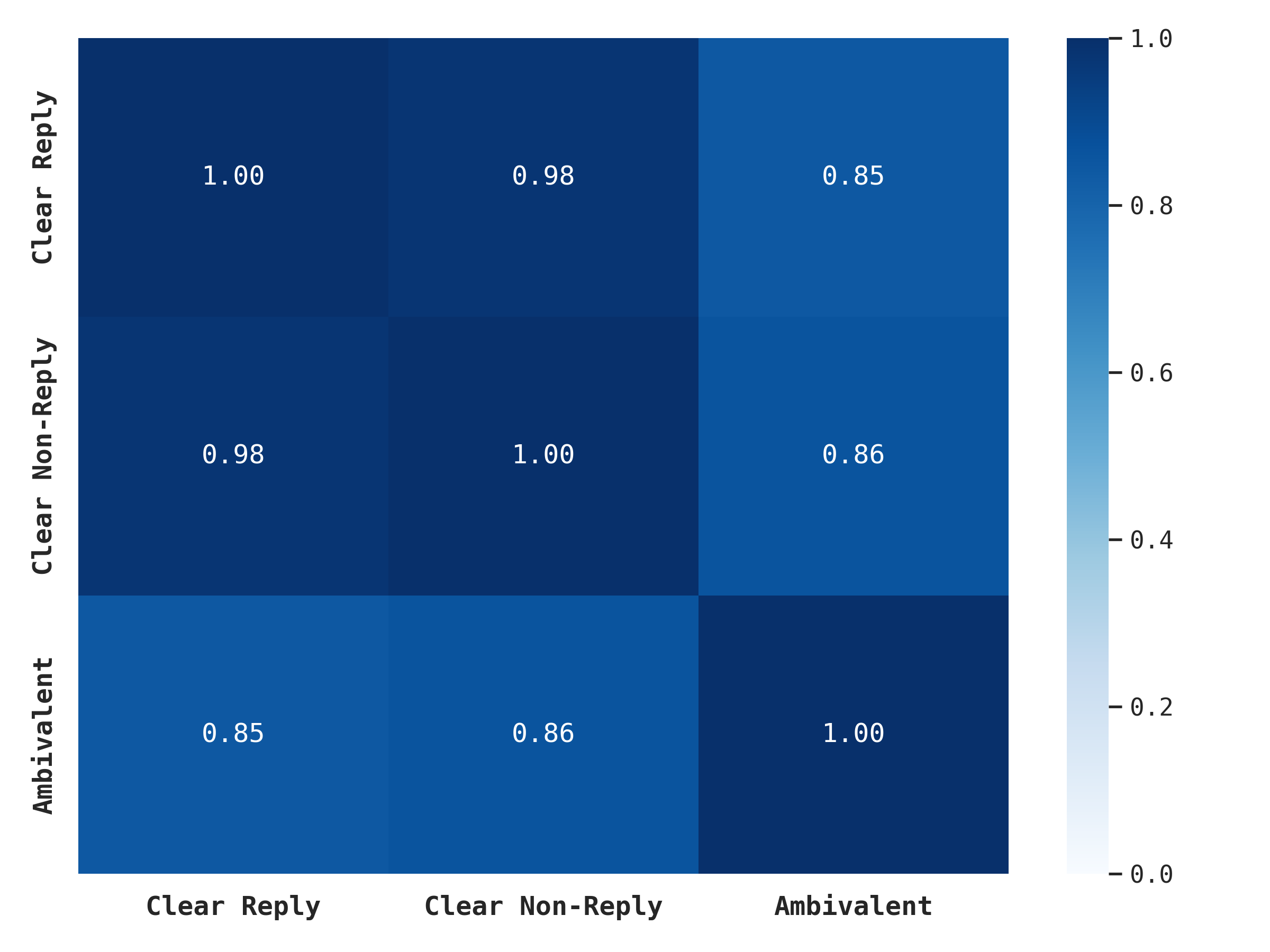}
\caption{Pairwise Fleiss' $\kappa$ between clarity categories.}
\label{fig:pairwise_kappa_clarity}
\end{figure}

Figure~\ref{fig:pairwise_kappa_clarity} shows pairwise Fleiss' $\kappa$ across the three clarity labels. All pairwise values are at least 0.85, with the lowest agreement at the \textit{Clear Reply}-\textit{Ambivalent} boundary ($\kappa=0.85$). Disagreement is mainly found between \textit{Clear Reply} and \textit{Ambivalent} answers rather than between the two non-ambiguous labels (\textit{Clear Reply} vs.\ \textit{Clear Non-Reply}: $\kappa=0.98$).

\begin{figure}[ht]
\centering
\includegraphics[width=0.95\columnwidth]{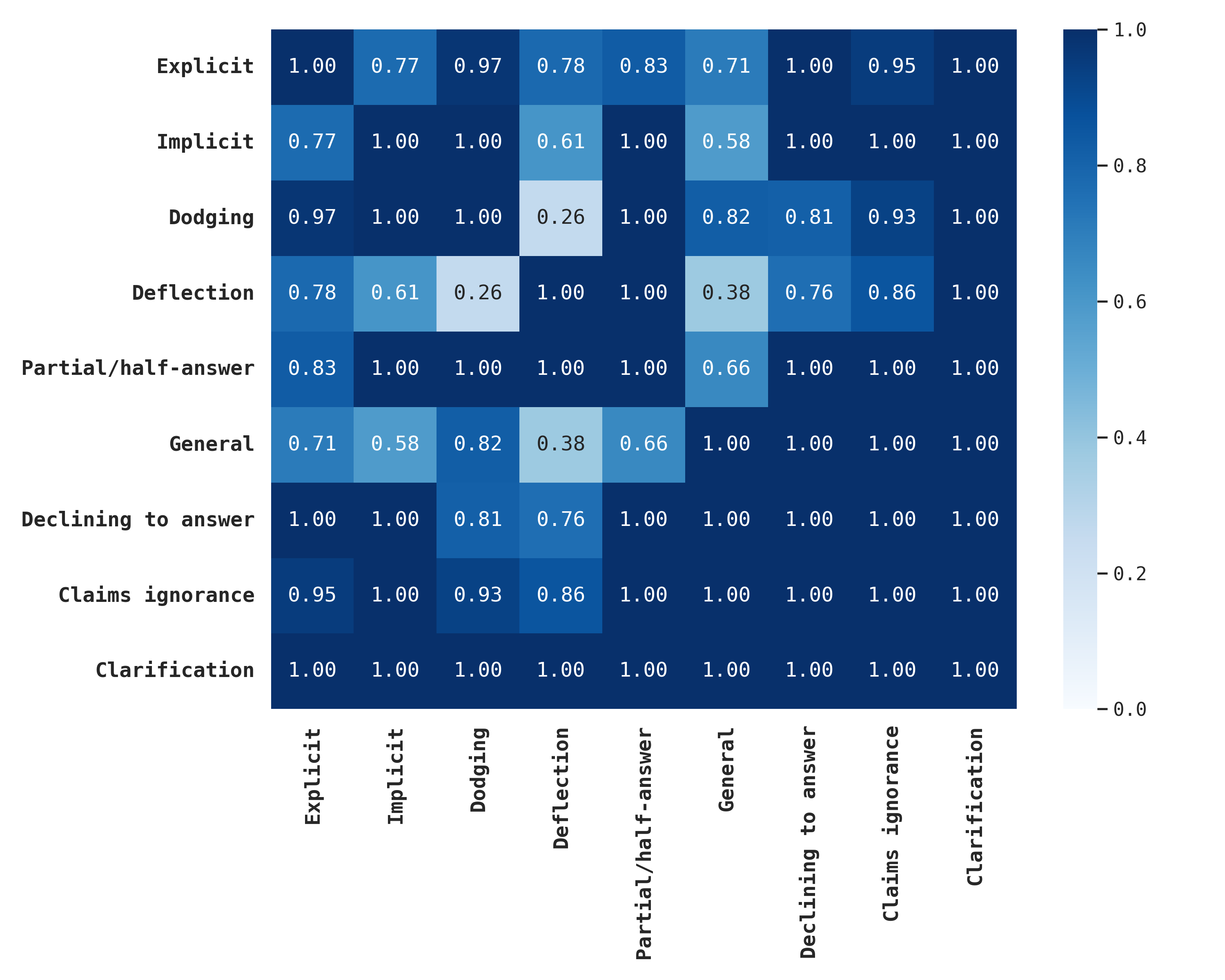}
\caption{Pairwise Fleiss' $\kappa$ between evasion categories.}
\label{fig:pairwise_kappa_evasion}
\end{figure}

Figure~\ref{fig:pairwise_kappa_evasion} shows the same matrix for the nine evasion categories. The lowest pairwise $\kappa$ values occur for \textit{Deflection}, in particular \textit{Deflection}-\textit{Dodging} ($\kappa=0.26$) and \textit{Deflection}-\textit{General} ($\kappa=0.38$). These confusions reflect how close the categories are in the taxonomy. \textit{Dodging} and \textit{Deflection} both involve leaving the question unanswered. They differ in whether the answer acknowledges the question before shifting away. \textit{General} and \textit{Deflection} differ in whether the answer remains on topic at all, which depends on how broadly the topic is interpreted.

\begin{figure}[ht]
\centering
\includegraphics[width=\columnwidth]{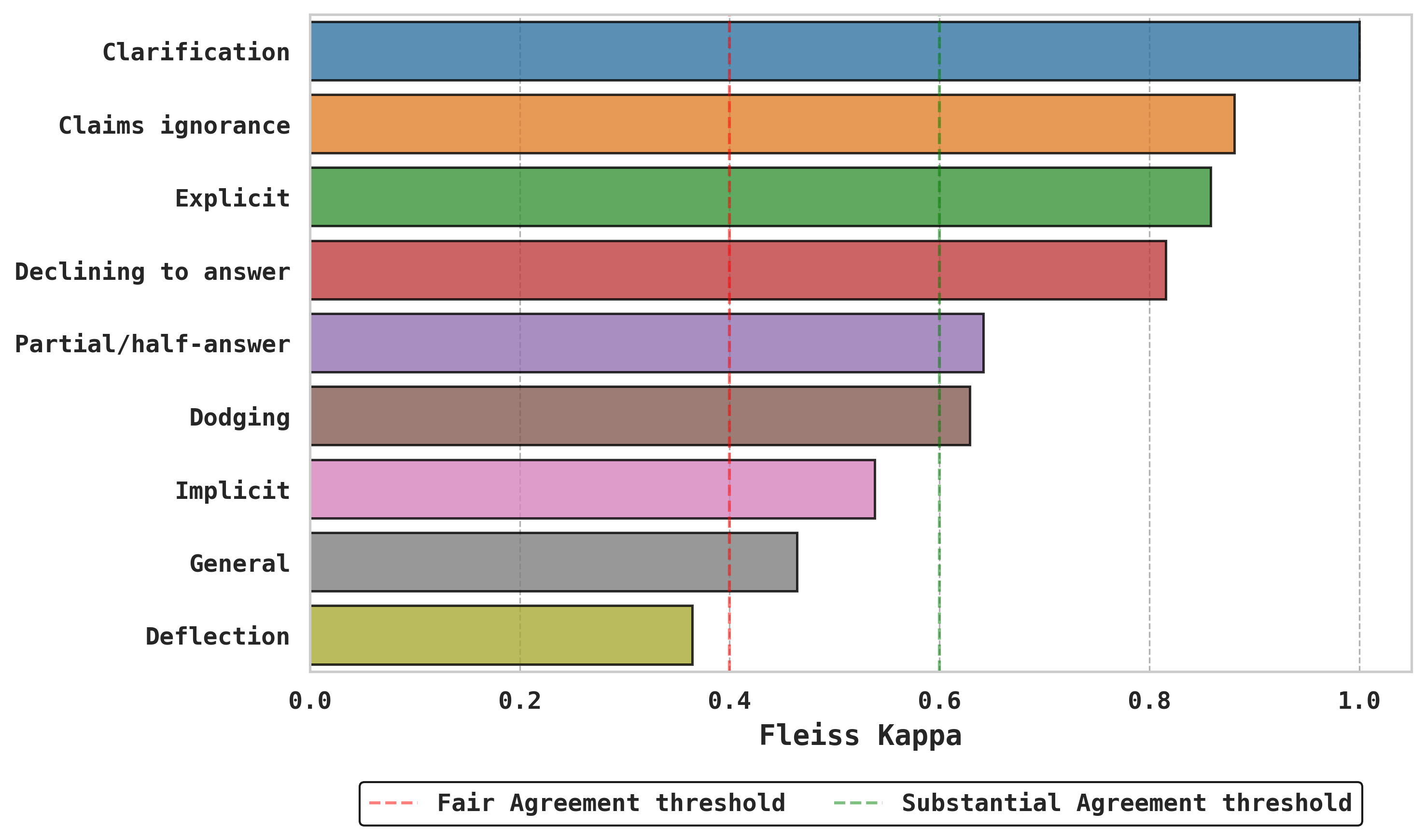}
\caption{One-vs-rest Fleiss' $\kappa$ per evasion category.}
\label{fig:category_kappas}
\end{figure}

Figure~\ref{fig:category_kappas} shows the per-category one-vs-rest Fleiss' $\kappa$. Six categories cross the substantial-agreement threshold of $\kappa=0.6$ (\textit{Clarification}, \textit{Claims ignorance}, \textit{Explicit}, \textit{Declining to answer}, \textit{Partial/half-answer}, \textit{Dodging}); two fall between fair and substantial (\textit{Implicit}, \textit{General}); one falls in the fair range (\textit{Deflection}).

\subsection{Expert Benchmark Evaluation}
\label{app:annotation:benchmark}

\begin{figure}[ht]
\centering
\includegraphics[width=\columnwidth]{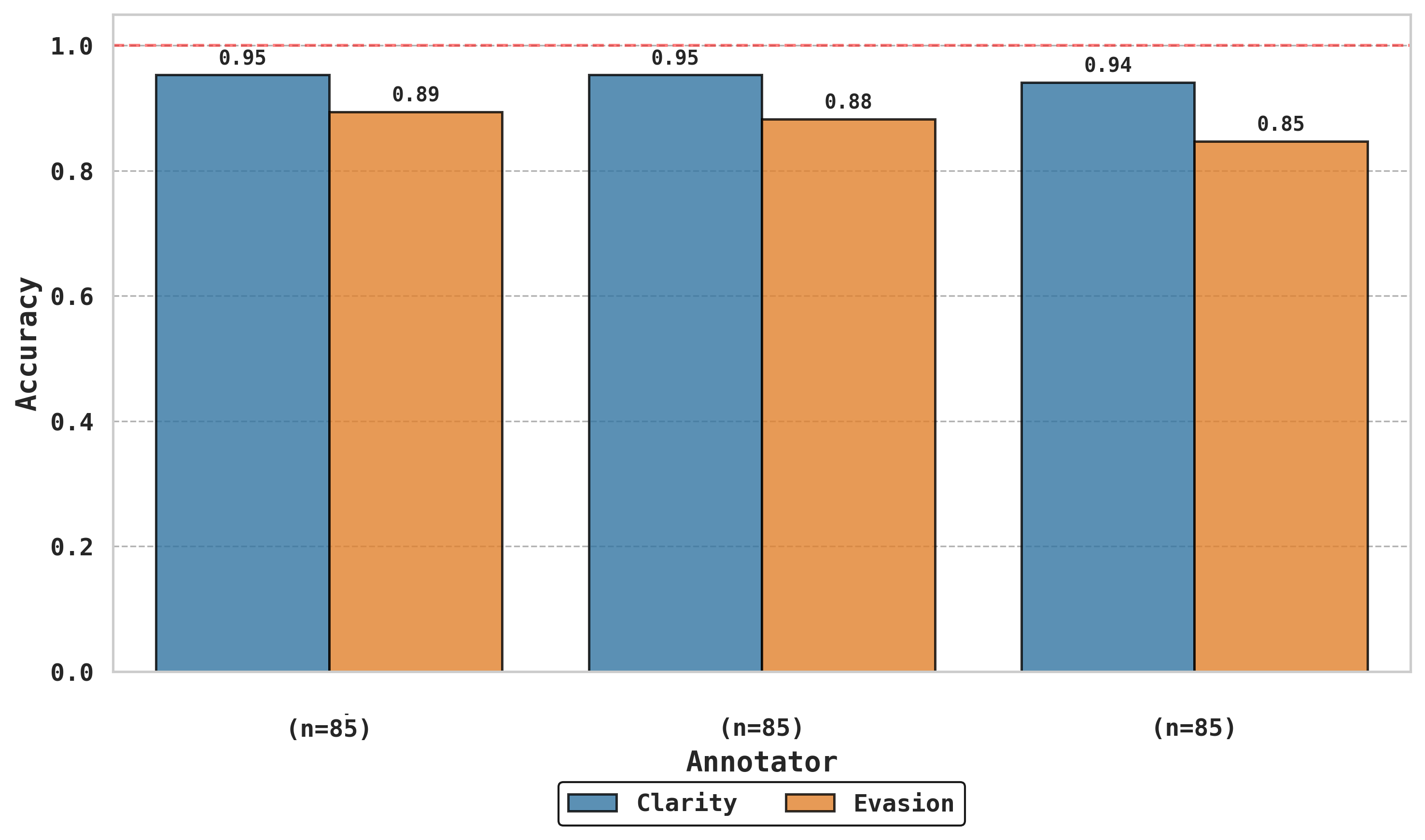}
\caption{Per-annotator clarity and evasion accuracy against expert benchmark.}
\label{fig:benchmark_accuracy}
\end{figure}

The linguistic expert independently labeled a benchmark of 85 instances drawn from the test split. Figure~\ref{fig:benchmark_accuracy} reports per-annotator accuracy against the expert: 95.3\%, 95.3\%, 94.1\% on clarity and 89.4\%, 88.2\%, 84.7\% on evasion. Clarity accuracy is 5--10 percentage points higher than evasion accuracy for each annotator, which is consistent with the greater difficulty of the nine-class task compared to the three-class task.

\section{Exploratory Data Analysis}
\label{app:eda}

This appendix reports descriptive statistics over the $3{,}574$ question-answer pairs in the dataset ($3{,}278$ train; $296$ test). The statistics are computed over the full dataset and use the majority-vote labels on the test set for both clarity and evasion categories.

\subsection{Label Distribution}
\label{sec:eda:labels}

\begin{figure}[ht]
\centering
\includegraphics[width=\linewidth]{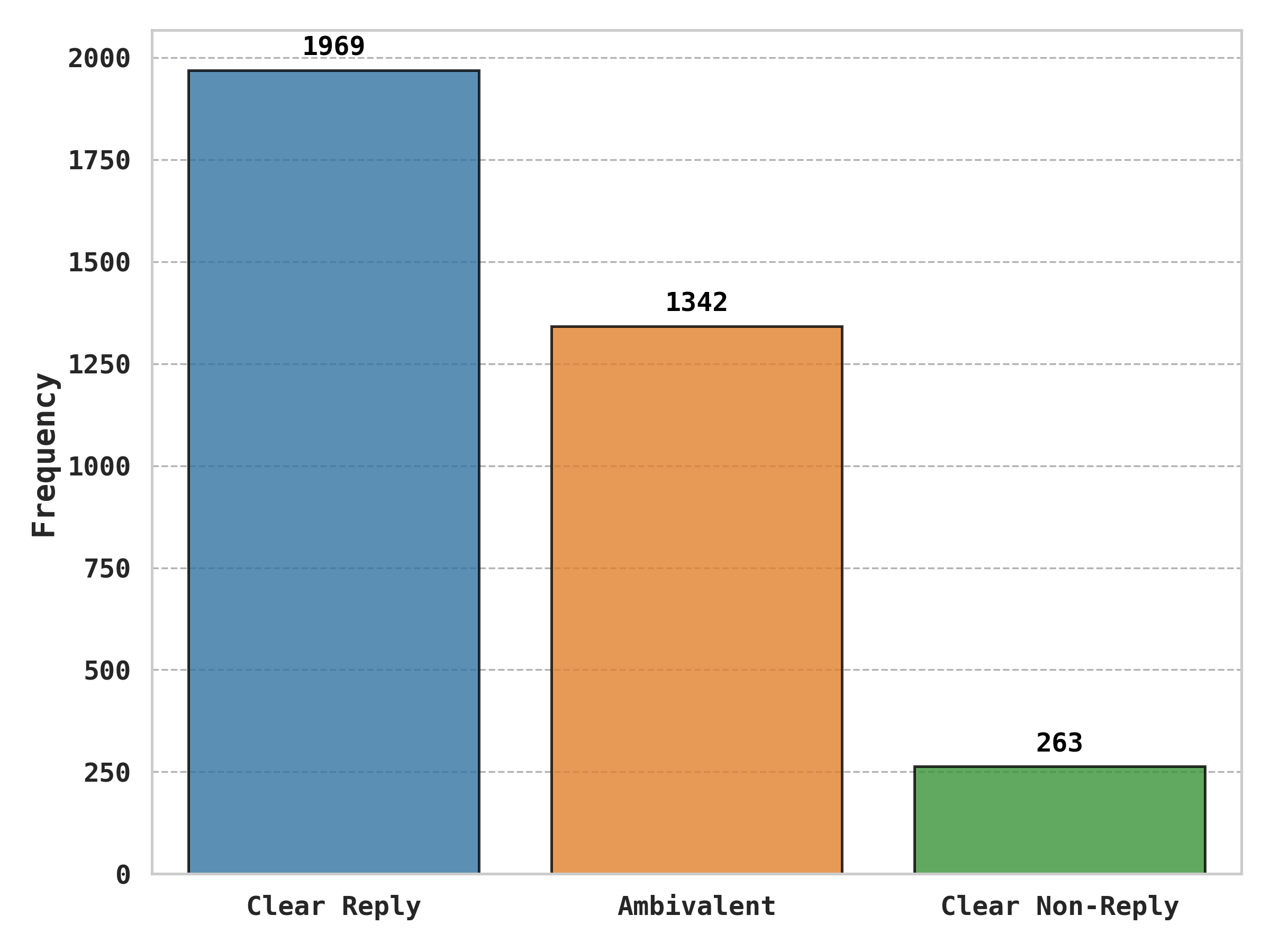}
\caption{Overall label distribution for Level 1 (Clarity).}
\label{fig:eda_clarity_dist}
\end{figure}

\begin{figure}[ht]
\centering
\includegraphics[width=\linewidth]{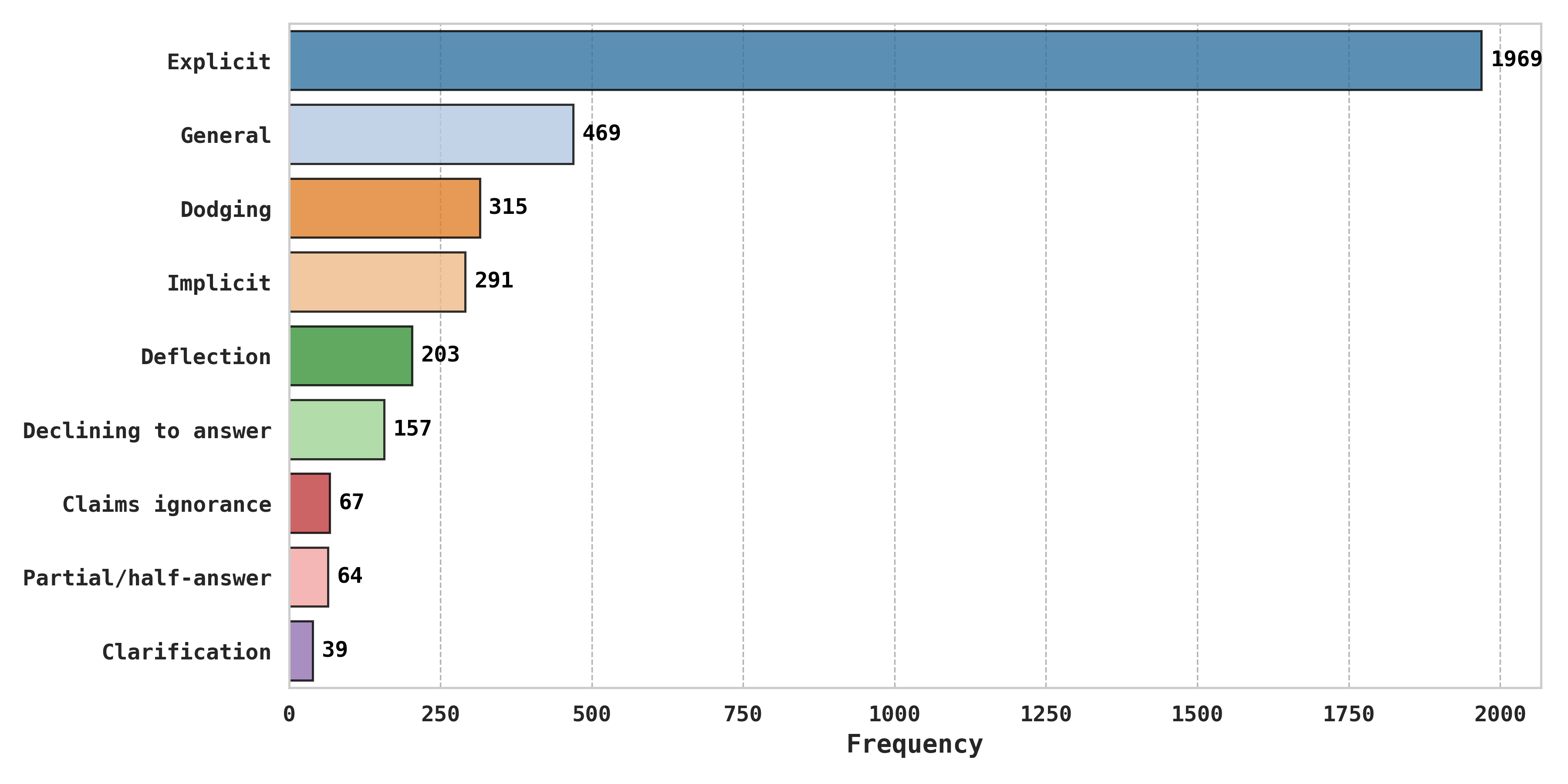}
\caption{Overall label distribution for Level 2 (Evasion).}
\label{fig:eda_evasion_dist}
\end{figure}

Figures~\ref{fig:eda_clarity_dist} and~\ref{fig:eda_evasion_dist} show the label distributions for clarity and evasion. \textit{Explicit} answers make up the majority of the dataset (55.09\%). The other eight evasion strategies are much less common, with \textit{Clarification} occurring in only 1.09\% of the question-answer pairs.

\begin{figure}[ht]
\centering
\includegraphics[width=\linewidth]{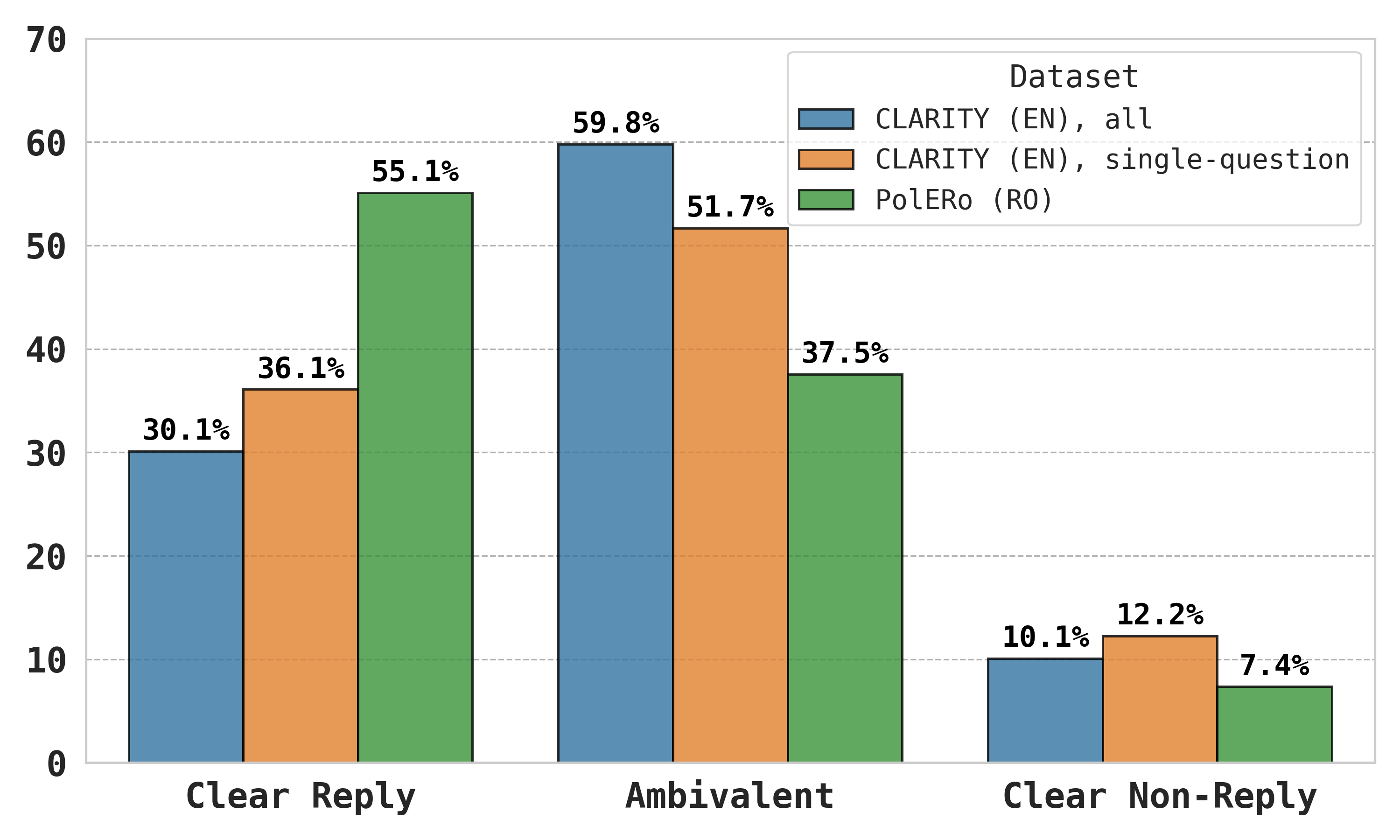}
\caption{Clarity distribution of the English CLARITY corpus before and after restricting
it to single-question instances, compared with PolERo.}
\label{fig:eda_fragmentation}
\end{figure}

CLARITY splits multi-part interviewer turns into one labeled row per sub-question using ChatGPT \cite{thomas2024}, whereas PolERo removes them entirely. In CLARITY, $67.4\%$ of instances are such fragments, and they are labeled \textit{Ambivalent} far more often than single-question instances ($63.7\%$ vs.\ $51.7\%$). When limiting \textsc{clarity} to examples with a single question ($N = 1{,}225$): \textit{Ambivalent} falls from $59.8\%$ to $51.7\%$ and \textit{Clear Reply} rises from $30.1\%$ to $36.1\%$ (Figure~\ref{fig:eda_fragmentation}). The difference between the two distributions shrinks minimally.

\subsection{Distribution by Speaker}
\label{sec:eda:president}

\begin{figure}[ht]
\centering
\includegraphics[width=\linewidth]{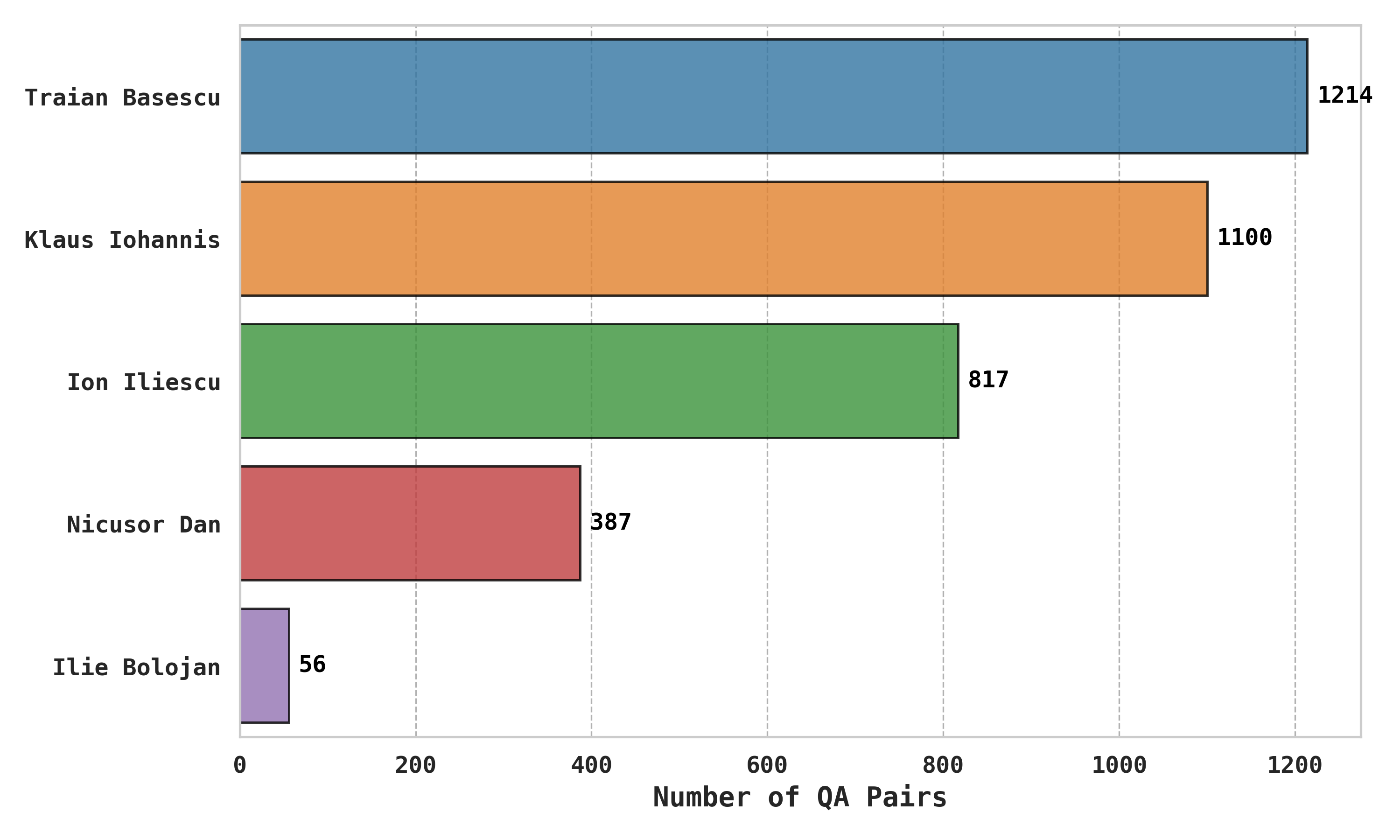} 
\caption{Total number of question-answer pairs per speaker.}
\label{fig:eda_president_volume}
\end{figure}

The corpus covers five Romanian presidents (Figure~\ref{fig:eda_president_volume}). Coverage reflects data availability: Băsescu ($1{,}214$) and Iohannis ($1{,}100$) account for 64.8\% of the corpus; Bolojan ($56$) is much less frequent, as he has been an interim president for 103 days. 

\begin{figure}[ht]
\centering
\includegraphics[width=\linewidth]{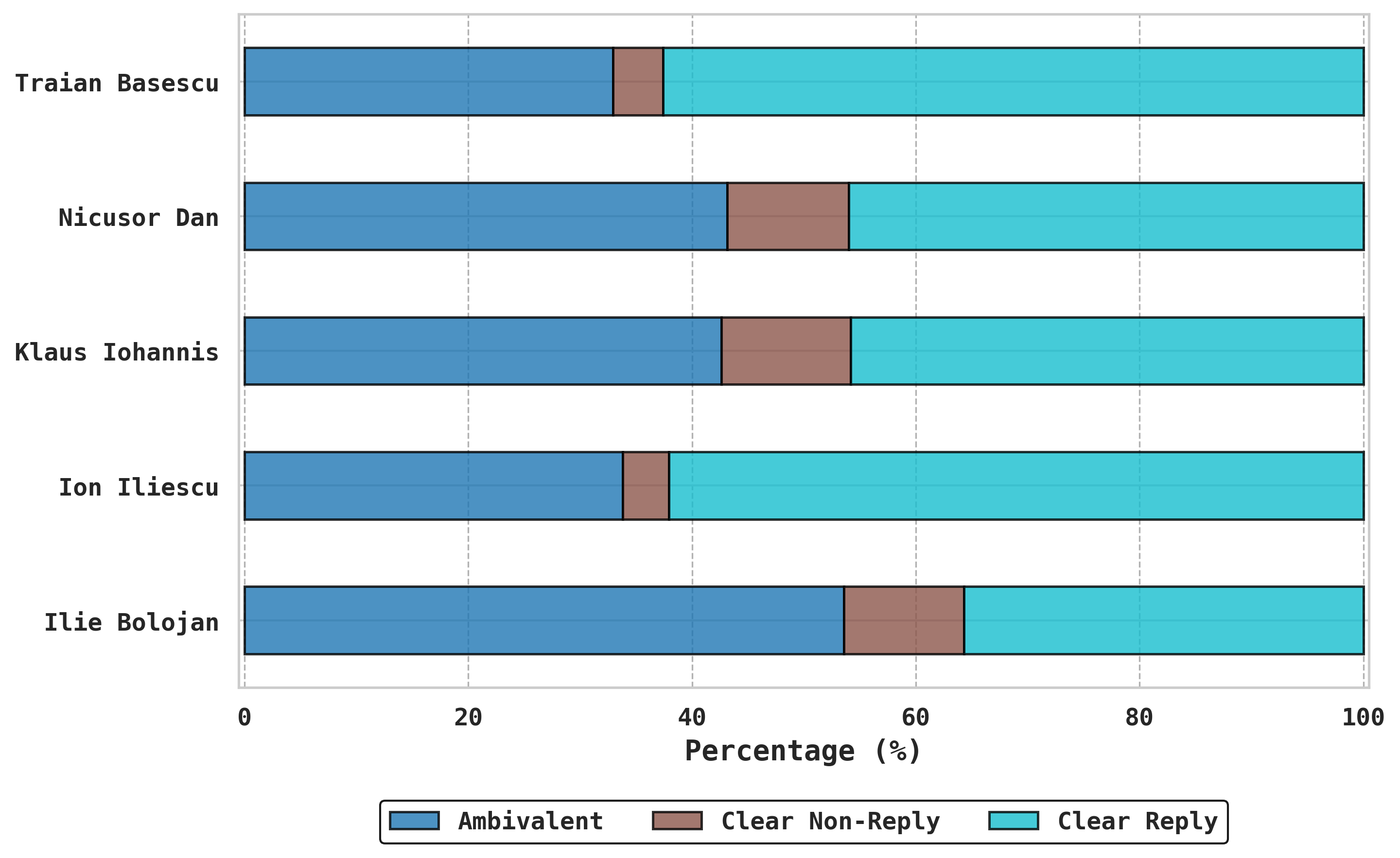} 
\caption{Clarity label distribution per speaker.}
\label{fig:eda_president_clarity}
\end{figure}

Figure~\ref{fig:eda_president_clarity} shows the clarity rates for each speaker. Iliescu and Băsescu give the most direct answers, with \textit{Clear Reply} rates of 62.06\% and 62.60\%. In contrast, Bolojan gives the most \textit{Ambivalent} answers (53.57\%), and Iohannis has the highest rate of \textit{Clear Non-Replies} (11.55\%).

\begin{figure}[ht]
\centering
\includegraphics[width=\columnwidth]{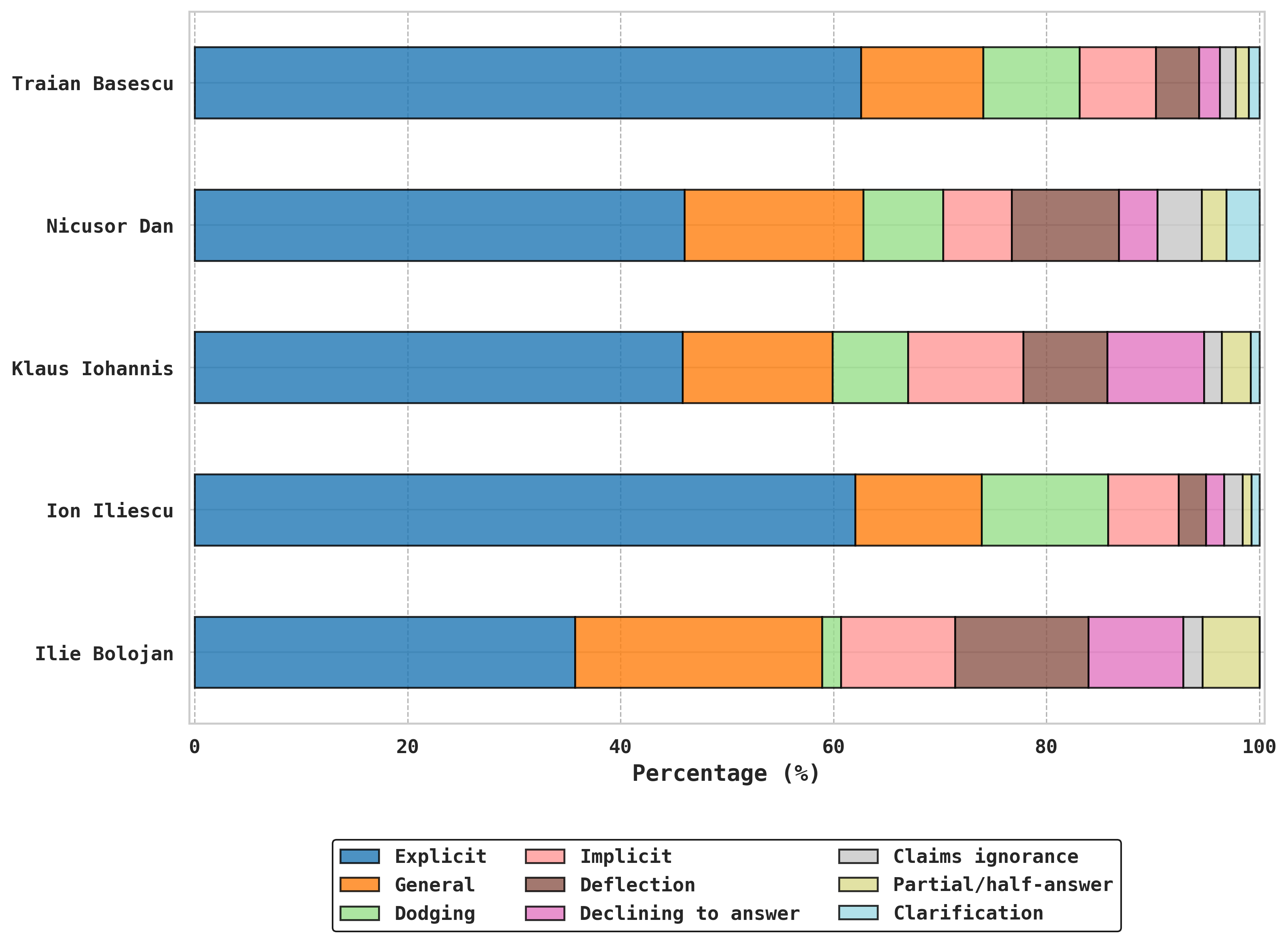}
\caption{Evasion label distribution per speaker.}
\label{fig:eda_president}
\end{figure}

Evasion distributions per speaker are shown in Figure~\ref{fig:eda_president}. \textit{Dodging} is most frequent for Iliescu (11.87\%) and Băsescu (9.06\%). \textit{Deflection} is most common for Bolojan (12.50\%) and Dan (10.08\%). \textit{Declining to answer} is highest for Iohannis (9.09\%) and Bolojan (8.93\%). The post-2014 speakers (Iohannis, Bolojan, and Dan) show a more balanced distribution of evasion types, with higher proportions of \textit{General}, \textit{Deflection}, and \textit{Declining to answer} responses. In contrast, Iliescu and Băsescu rely more heavily on \textit{Dodging}, while still remaining largely dominated by \textit{Explicit} answers.

\subsection{Distribution by Communicative Context}
\label{sec:eda:context}

We assigned a context label to each source transcript from event titles using keywords associated with each category. Interviews make up the majority of the dataset (1,904 pairs), followed by declarations (1,213 pairs) and press conferences (457 pairs).

\begin{figure}[ht]
\centering
\includegraphics[width=\linewidth]{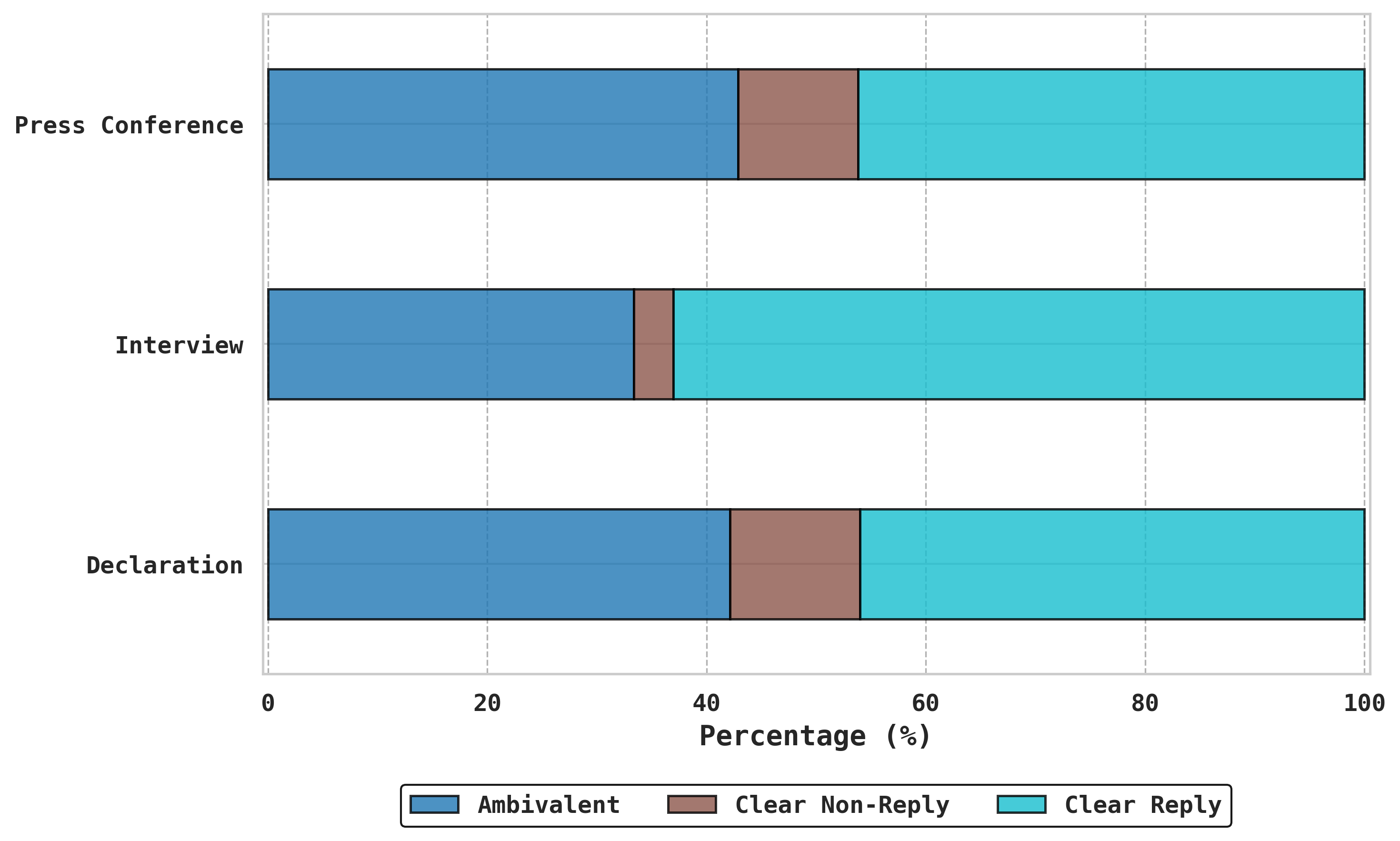}
\caption{Clarity distribution per communicative context.}
\label{fig:eda_context_clarity}
\end{figure}

Figure~\ref{fig:eda_context_clarity} reports clarity rates across the three resulting contexts. Interviews have the highest rate of \textit{Clear Replies} (63.03\%) and the lowest rate of \textit{Clear Non-Replies} (3.62\%). Declarations and press conferences show very similar clarity distributions.

Part of the higher clarity rate in interviews is linked to differences in topic distribution. \textit{Personal \& Others} makes up 38.39\% of interview questions, compared to only 12.94\% in declarations and 8.75\% in press conferences (see Section \ref{sec:eda:category}). This category also has the highest \textit{Clear Reply} rate of any topic (65.62\%). Press conferences contain more \textit{Justice \& Anti-corruption} and \textit{Domestic Policy} questions, which have the lowest \textit{Clear Reply} rates.

\begin{figure}[ht]
\centering
\includegraphics[width=\columnwidth]{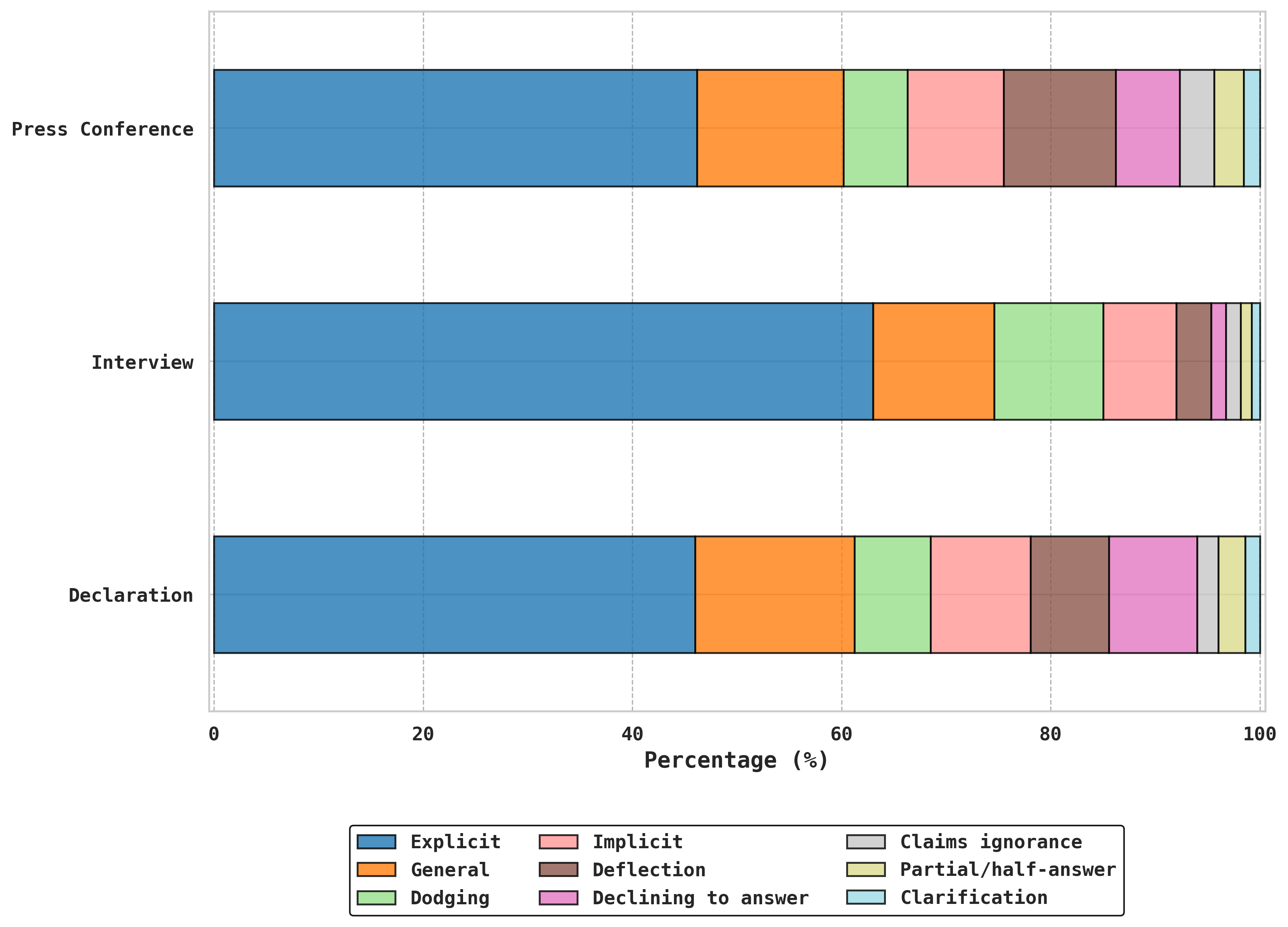}
\caption{Evasion label distribution per communicative context.}
\label{fig:eda_context}
\end{figure}

Evasion distributions per context are shown in Figure~\ref{fig:eda_context}. Interviews show a higher percentage of \textit{Dodging} (10.45\%), but low levels of the other evasion types. Declarations and press conferences have roughly four to six times the rate of \textit{Declining to answer} (8.41\% and 6.13\%) and higher \textit{Deflection} (7.50\% and 10.72\%) compared to interviews (1.42\% and 3.31\%). Press conferences have the highest \textit{Deflection} rate overall. If we exclude interviews and consider only the more formal settings of declarations and press conferences ($n{=}1{,}670$), the overall proportion of direct answers drops significantly: the \textit{Explicit} rate falls to 46.05\%, while evasive categories such as \textit{Deflection} (8.38\%) and \textit{Declining to answer} (7.79\%) account for a larger share of responses.

\begin{figure}[ht]
\centering
\includegraphics[width=\columnwidth]{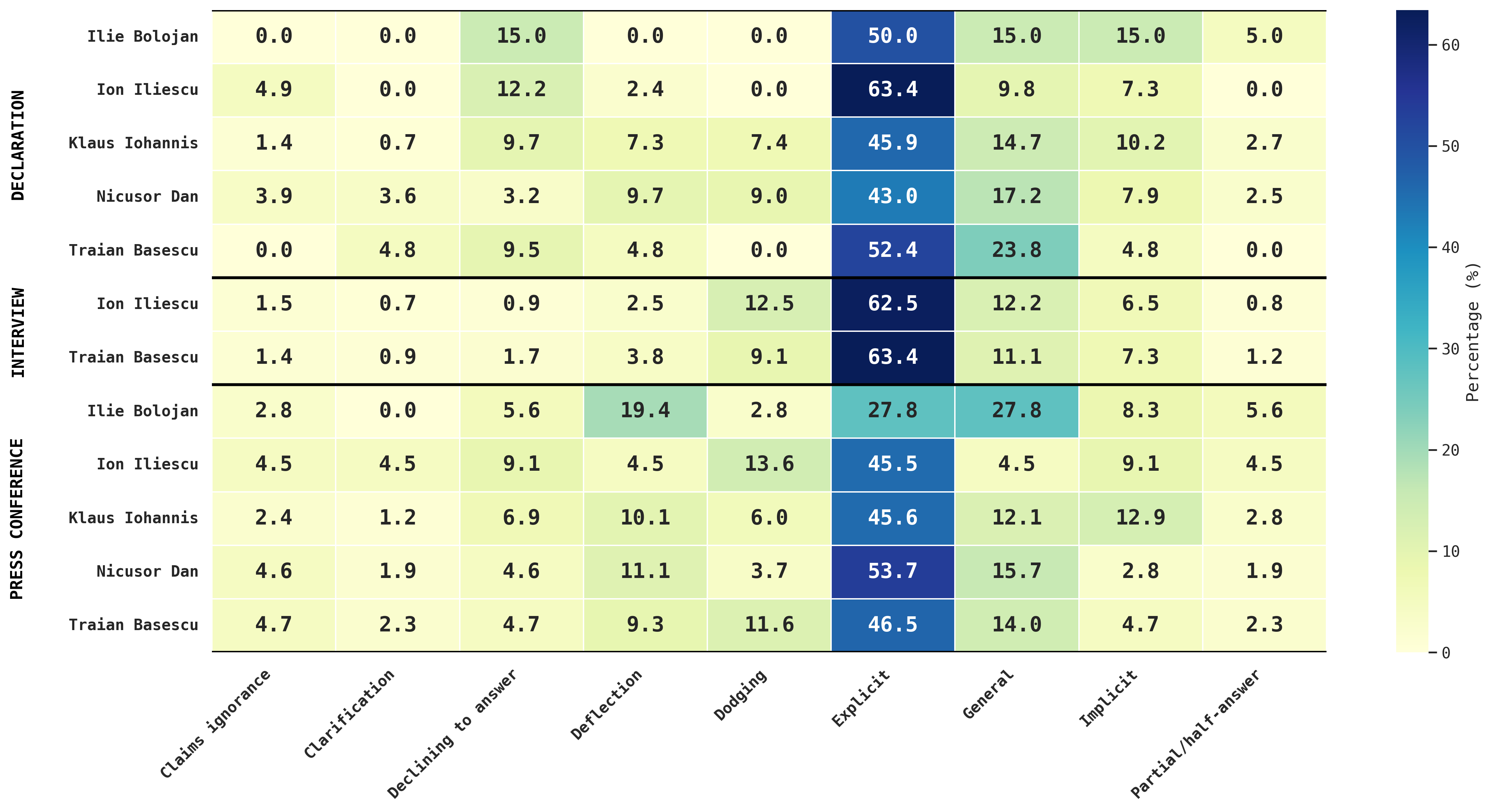}
\caption{Evasion distribution per speaker $\times$ context cell.}
\label{fig:eda_context_matrix}
\end{figure}

Figure~\ref{fig:eda_context_matrix} crosses speaker with context. Iliescu and Băsescu interviews ($n{=}754$, $n{=}1{,}150$) explain the high interview \textit{Explicit} rate (62.47\%, 63.39\%) and also account for most \textit{Dodging} (12.47\%, 9.13\%) in interviews. Iohannis declarations ($n{=}852$) account for the bulk of \textit{Declining to answer} in the declaration row (9.74\%). Băsescu declarations show the highest \textit{General} share (23.80\%) and Bolojan press conferences ($n{=}36$) the highest combined non-\textit{Explicit} share (72.22\%).

\subsection{Distribution by Topic Category}
\label{sec:eda:category}

\begin{figure}[ht]
\centering
\includegraphics[width=\linewidth]{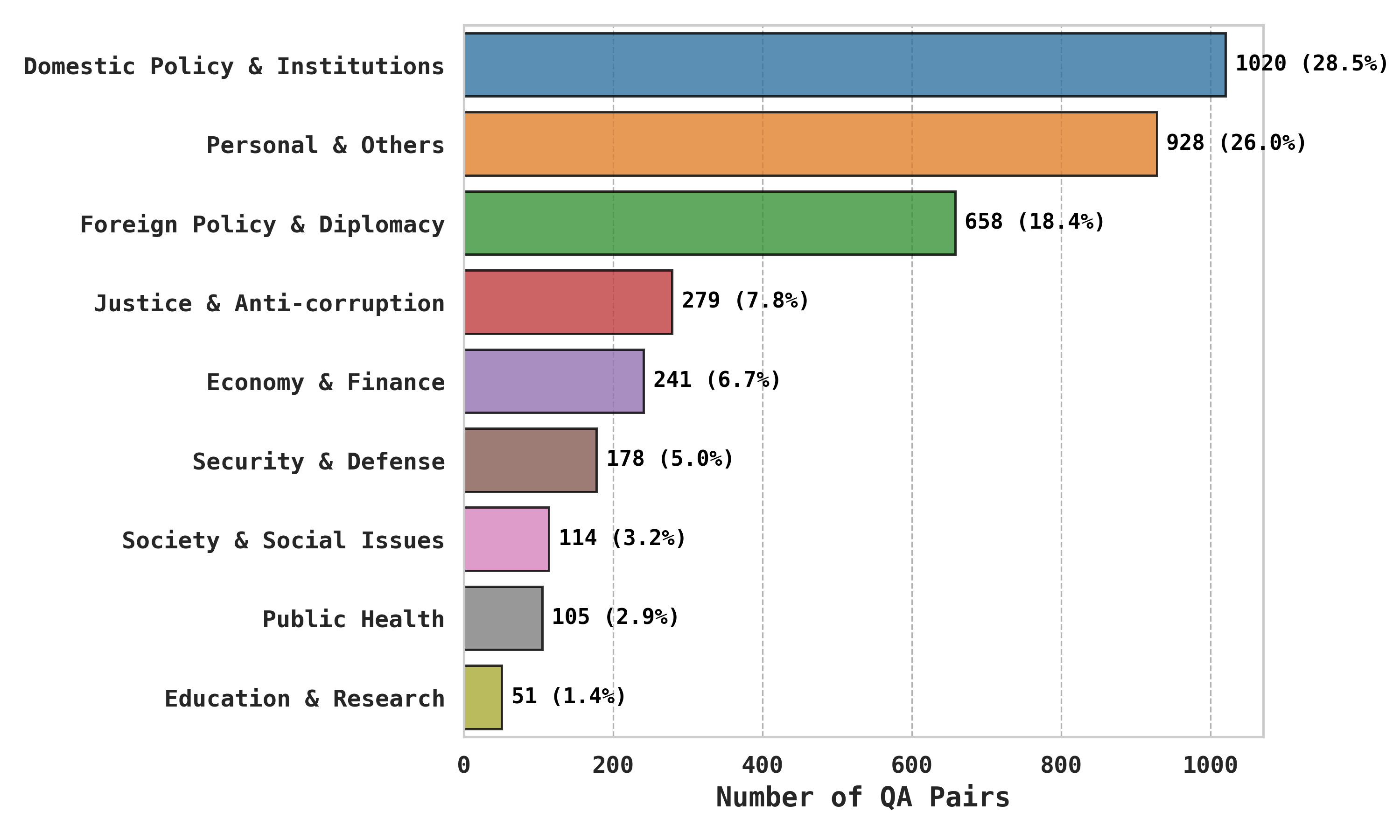} 
\caption{Total number of question-answer pairs per topic category.}
\label{fig:eda_category_volume}
\end{figure}

Categories were defined by first inspecting the question content and identifying nine main topics. GPT-5.4 then assigned a single category label to each question. Figure~\ref{fig:eda_category_volume} shows the overall distribution of these topics. 

\begin{figure}[ht]
\centering
\includegraphics[width=\linewidth]{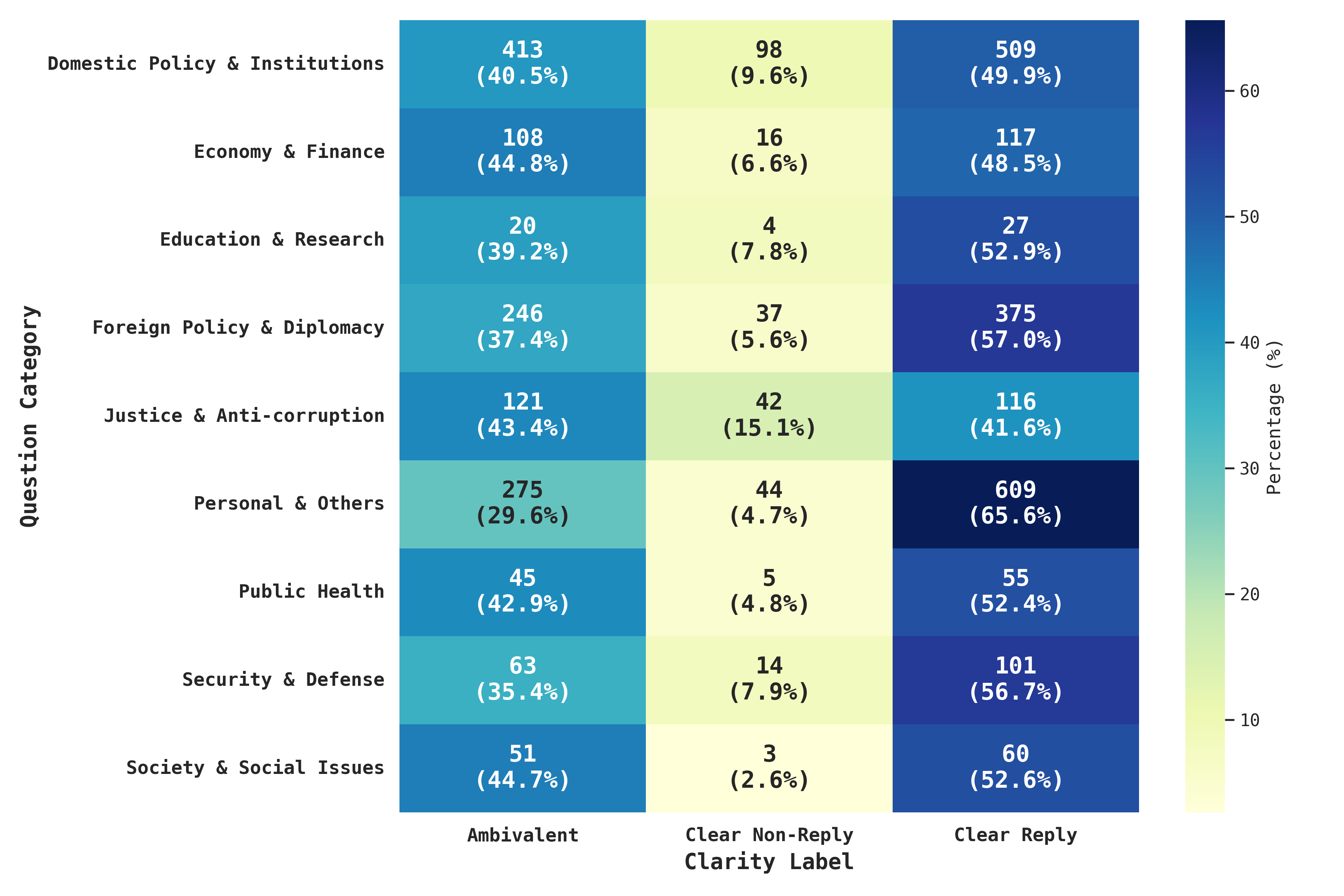} 
\caption{Clarity label distribution per topic category.}
\label{fig:eda_category_clarity}
\end{figure}

Figure~\ref{fig:eda_category_clarity} shows the clarity rates for each topic. Even though \textit{Justice \& Anti-corruption} questions make up only 7.81\% of the data, they trigger the highest rate of \textit{Clear Non-Replies} (15.05\%). 

\begin{figure}[ht]
\centering
\includegraphics[width=\columnwidth]{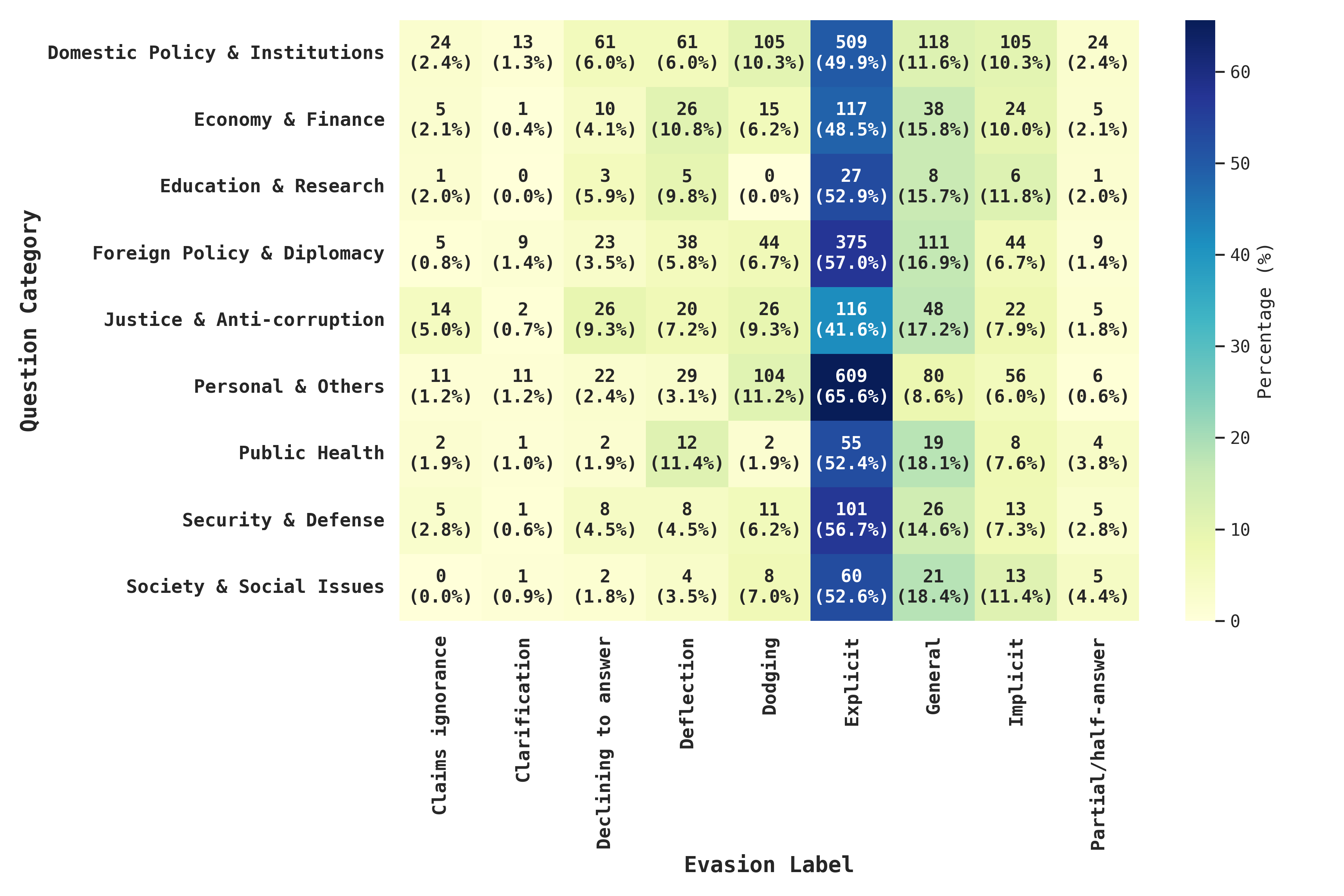}
\caption{Evasion label distribution per topic category.}
\label{fig:eda_category}
\end{figure}

Per-category evasion distributions are shown in Figure~\ref{fig:eda_category}. \textit{Personal \& Others} has the highest \textit{Dodging} share (11.21\%), although most responses in this category are still \textit{Explicit}. \textit{Justice \& Anti-corruption} has the highest rates of \textit{Declining to answer} (9.32\%) and \textit{Claims ignorance} (5.02\%). \textit{Education \& Research} shows the highest share of \textit{Implicit} (11.76\%), followed by \textit{Society \& Social Issues} (11.40\%). \textit{General} responses are most frequent for \textit{Society \& Social Issues} (18.42\%) and \textit{Public Health} (18.10\%).

\subsection{Temporal Coverage}
\label{sec:eda:time}

\begin{figure}[ht]
\centering
\includegraphics[width=\columnwidth]{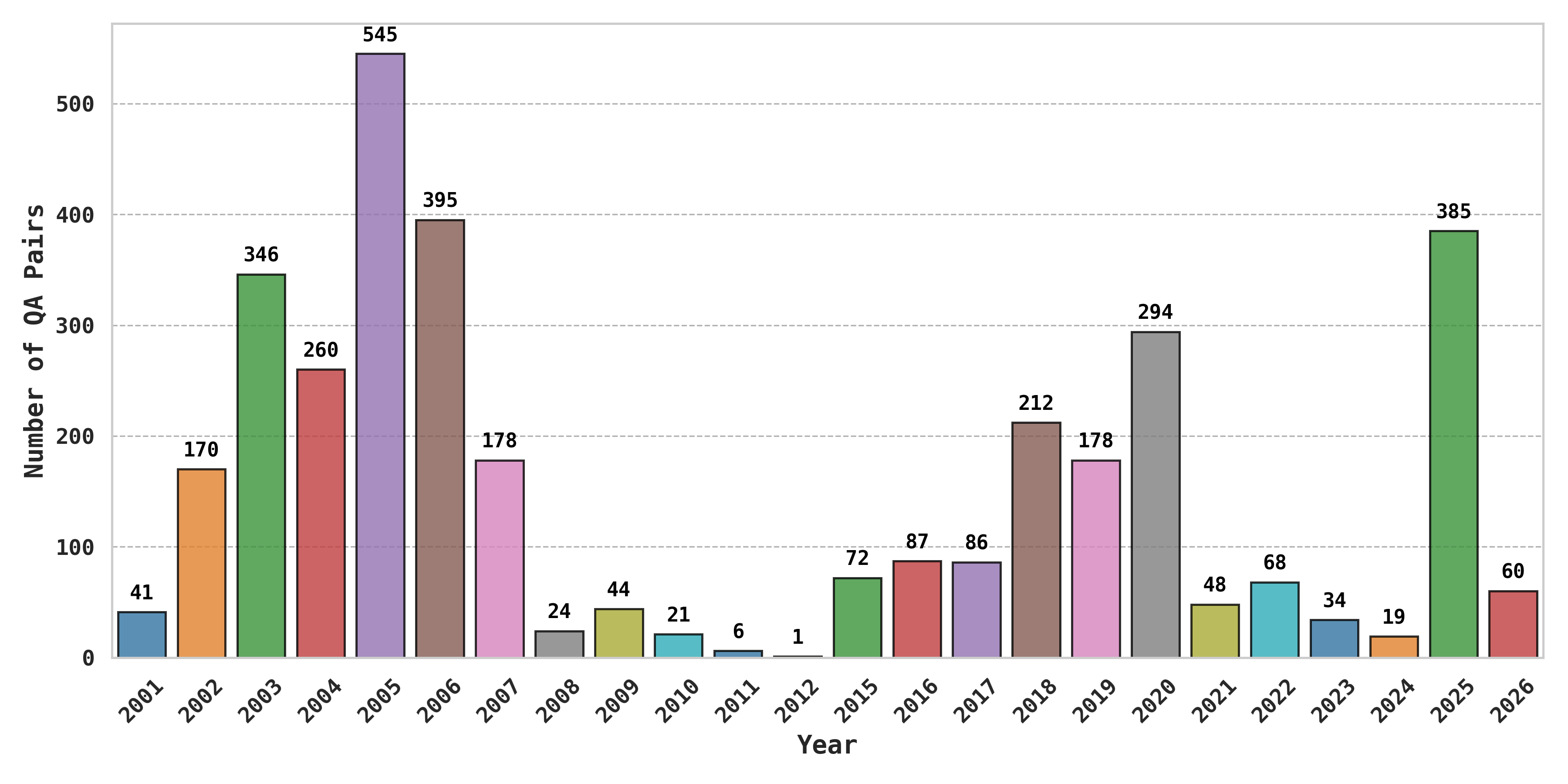}
\caption{Yearly volume of QA pairs, 2001--2026.}
\label{fig:eda_time_volume}
\end{figure}

\begin{figure}[ht]
\centering
\includegraphics[width=\columnwidth]{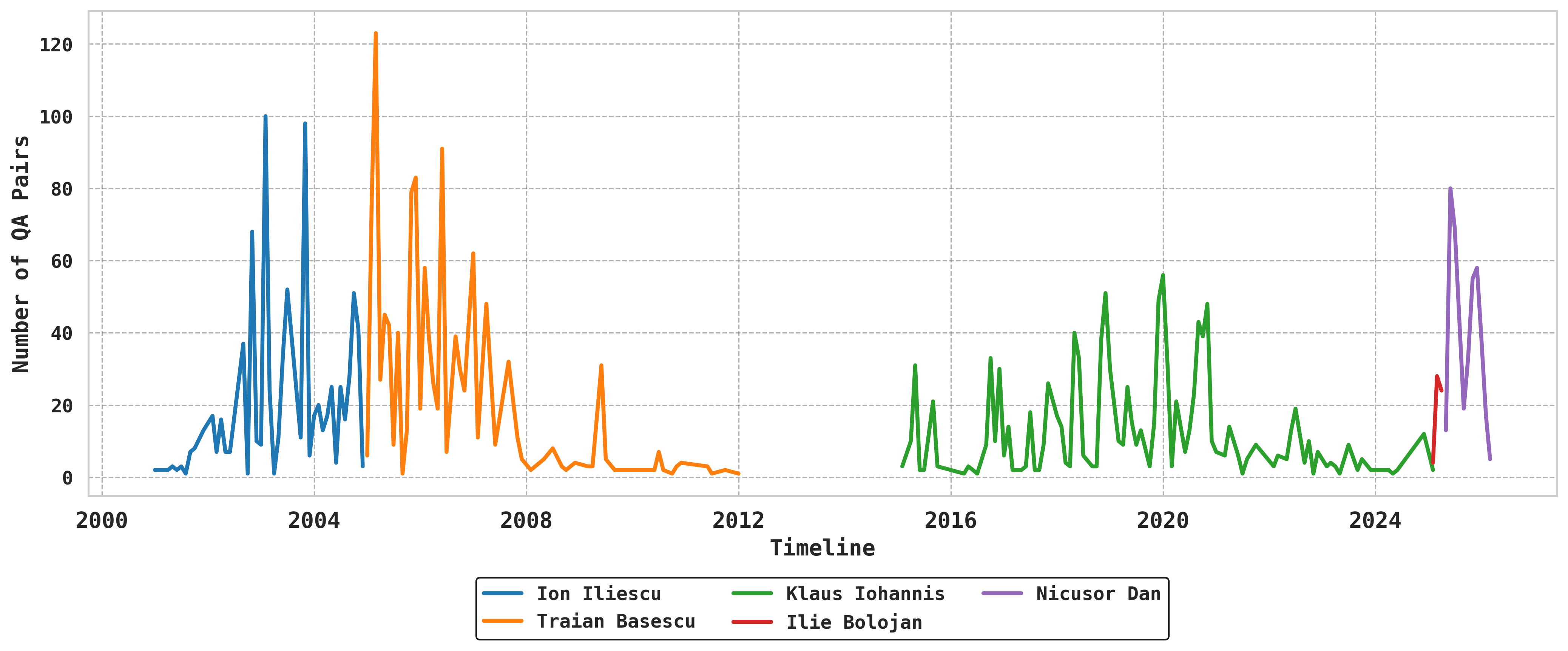}
\caption{Yearly QA volume per speaker.}
\label{fig:eda_time_president}
\end{figure}

QA pairs span $2001$--$2026$ in two main periods: $2001$--$2007$ (Iliescu and Băsescu, $1{,}935$ pairs) and $2015$--$2026$ (Iohannis, Bolojan, and Dan, $1{,}543$ pairs). The intervening $2008$--$2014$ window contains $96$ pairs, due to limited archival availability. Per-year volumes and per-speaker timelines are shown in Figures~\ref{fig:eda_time_volume} and~\ref{fig:eda_time_president}.

\begin{figure}[ht]
\centering
\includegraphics[width=\columnwidth]{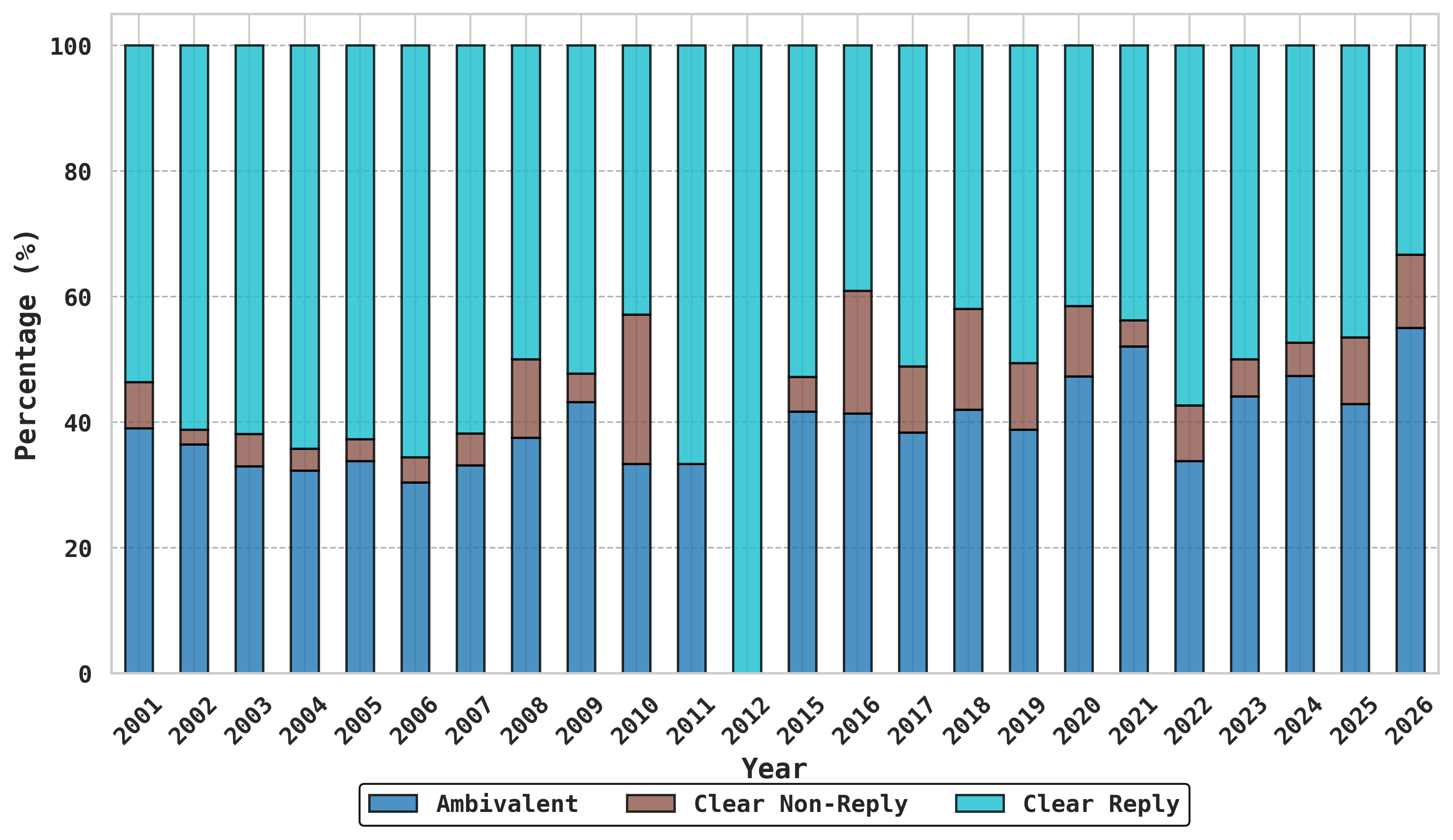}
\caption{Clarity label composition per year, 2001--2026.}
\label{fig:eda_time_clarity}
\end{figure}

Year-level clarity composition is shown in Figure~\ref{fig:eda_time_clarity}. \textit{Clear Non-Reply} rates remain below 6\% in 2001--2007, and increase to around 10--20\% in 2016--2026, with \textit{Ambivalent} responses increasing in parallel. 

\subsection{Length and Verbosity}
\label{sec:eda:length}

\begin{figure}[ht]
\centering
\includegraphics[width=\columnwidth]{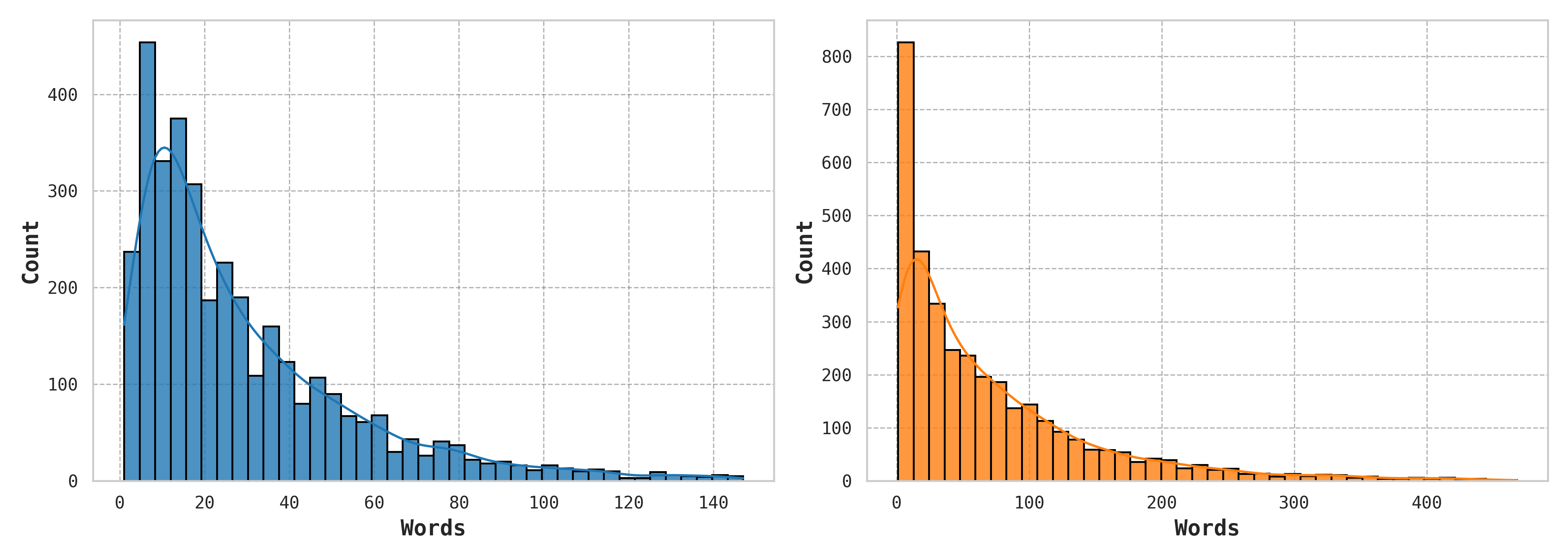}
\caption{Question and answer word-count distributions.}
\label{fig:eda_length}
\end{figure}

Mean question length is $30.7$ words, while the median is $21$ words; mean answer length is $76.9$ words, while the median is $45$ words. Overall length distributions are shown in Figure~\ref{fig:eda_length}.

\begin{figure}[ht]
\centering
\includegraphics[width=\linewidth]{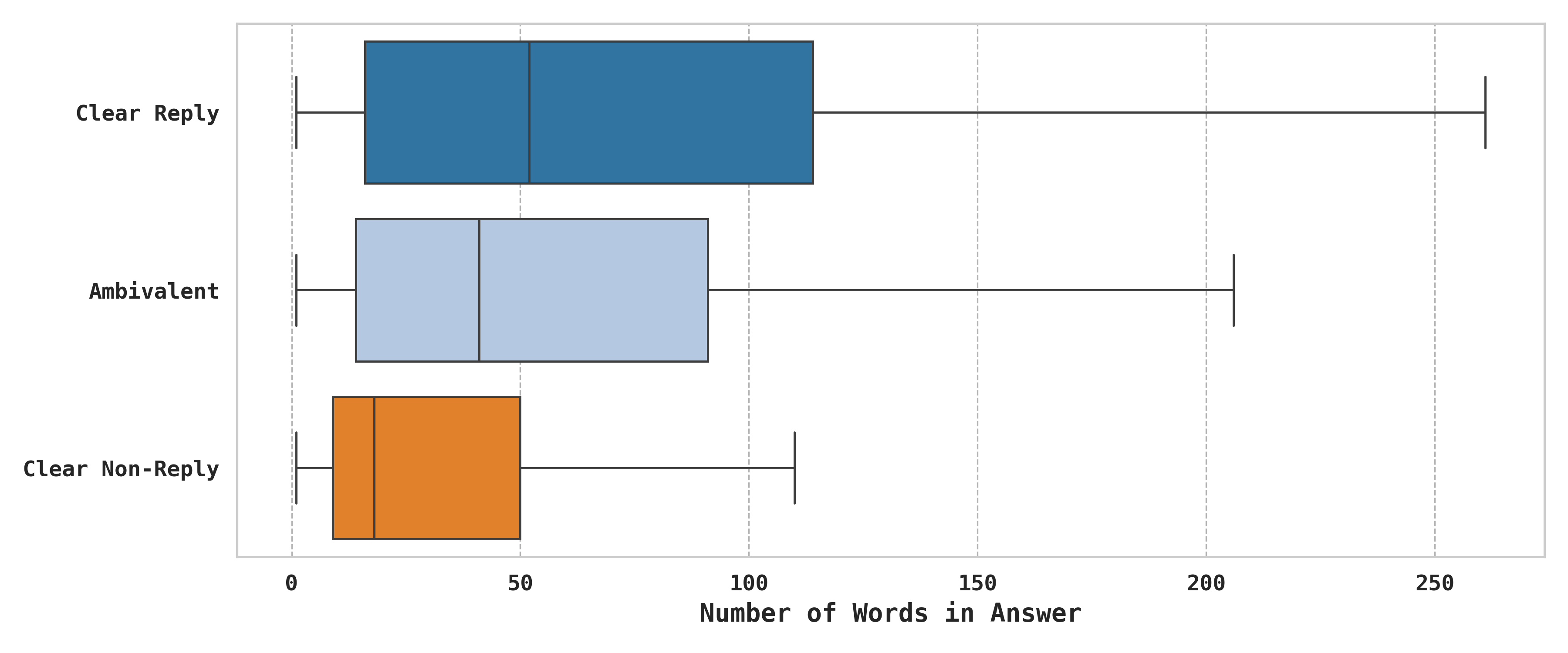} 
\caption{Answer word-count distribution per clarity label.}
\label{fig:eda_length_clarity}
\end{figure}

\begin{figure}[ht]
\centering
\includegraphics[width=\columnwidth]{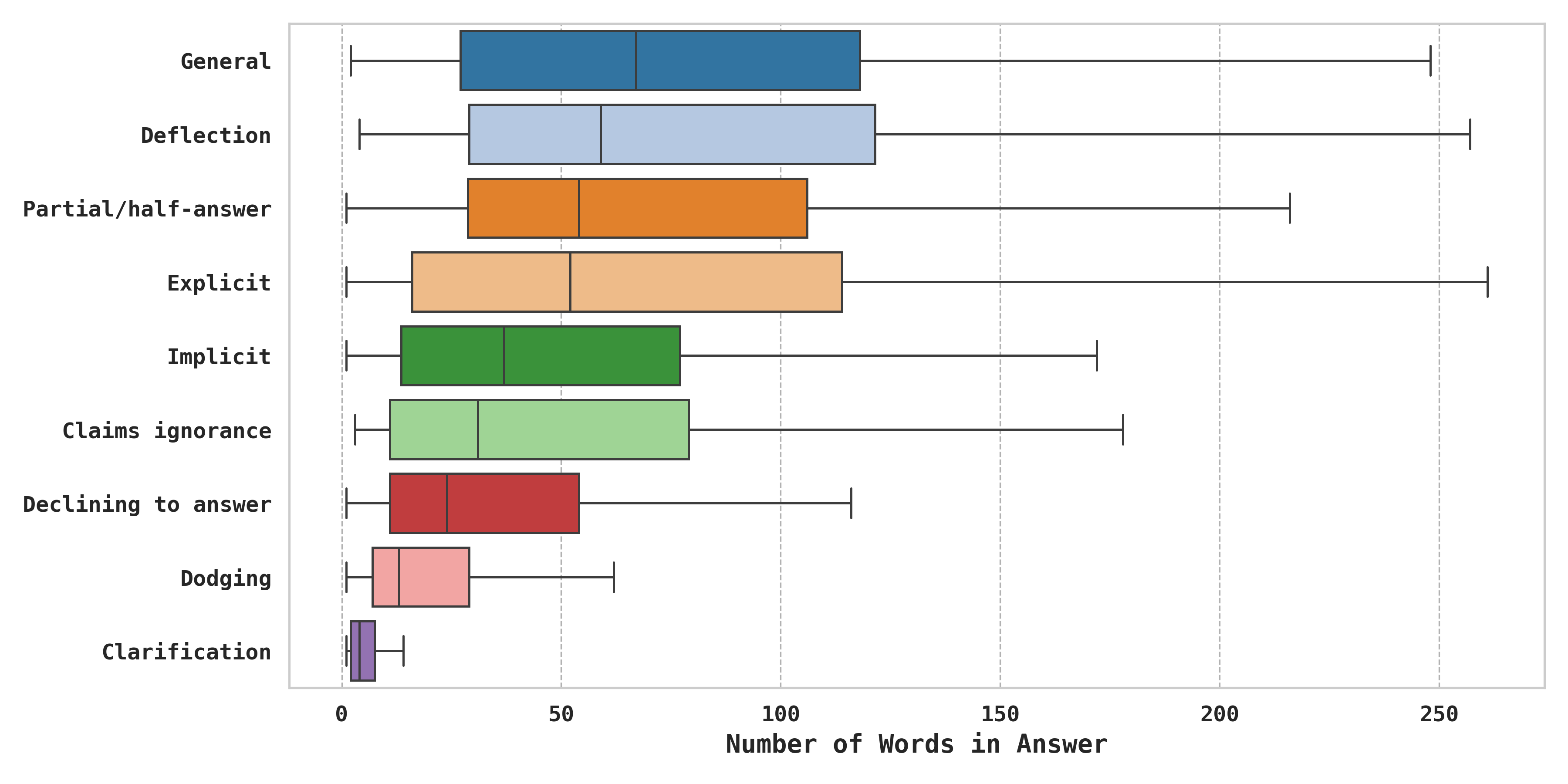}
\caption{Answer word-count distribution per evasion label.}
\label{fig:eda_length_evasion}
\end{figure}

Answer length differs substantially by label (Figures~\ref{fig:eda_length_clarity} and~\ref{fig:eda_length_evasion}). \textit{Clear Replies} average 88.4 words, \textit{Ambivalent} answers 67.4, and \textit{Clear Non-Replies} 39.7.

Looking at the specific evasion strategies, \textit{Clarification} answers are the shortest (mean $6.4$, median $4$) followed by \textit{Dodging} (mean $27.6$, median $13$). In contrast, \textit{General}, \textit{Deflection}, and \textit{Partial/half-answer} responses are the longest non-\textit{Explicit} categories (averaging $89.7$, $88.6$, and $87.3$ words). This aligns with verbose evasion tactics, where politicians surround the question with extra information instead of answering it directly.

\begin{figure}[ht]
\centering
\includegraphics[width=\columnwidth]{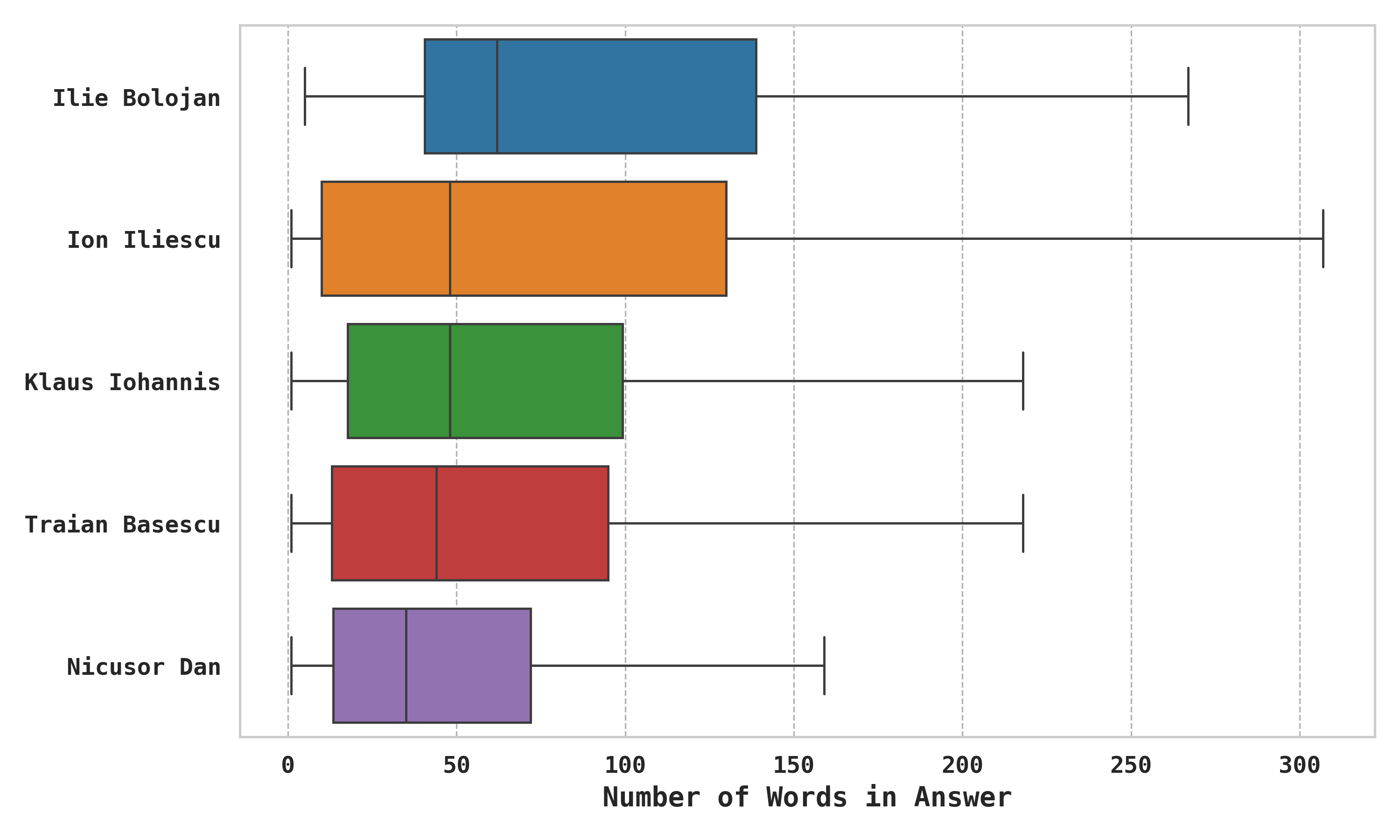}
\caption{Answer word-count distribution per speaker.}
\label{fig:eda_length_president}
\end{figure}

Per-speaker mean answer length ranges from $52.9$ words (Dan) to $110.6$ (Bolojan, $n{=}56$), with Iohannis at $73.1$, Băsescu at $69.3$ and Iliescu at $102.6$. The full per-speaker answer-length distribution is shown in Figure~\ref{fig:eda_length_president}. Speaker verbosity does not align with evasion: Iliescu is among the most verbose speakers and also one of the most \textit{Explicit}, while Dan is the most concise and is also predominantly \textit{Explicit}.

\section{Experiment Details}
\label{app:experiments}

This appendix expands the experimental setup and analysis summarised in Sections~\ref{sec:setup} and~\ref{sec:results}. 

\subsection{Evaluation}
\label{app:experiments:evaluation}

Across all experiments, we evaluate model performance using macro-averaged F1 as the primary metric. The macro-F1 is calculated as the unweighted mean of the class-specific F1 scores:

\begin{equation}
F1_{\text{macro}} = \frac{1}{|C|} \sum_{i=1}^{|C|} F1_i
\end{equation}

\noindent where $|C|$ is the total number of classes (3 for clarity, 9 for evasion), and $F1_i$ is the harmonic mean of precision ($P_i$) and recall ($R_i$) for class $i$:

\begin{equation}
F1_i = 2 \cdot \frac{P_i \cdot R_i}{P_i + R_i}
\end{equation}

To compute precision and recall, model predictions are compared against human annotations. 

For the 3-class clarity task, predictions are evaluated strictly against a single ground-truth label obtained via majority vote. A prediction is considered correct only if it exactly matches this label.

For the 9-class evasion task, a prediction is considered correct if it matches at least one annotator's label, which reflects valid differences in human interpretation. If it matches none, we use the majority-vote label as the ground truth for an incorrect prediction.

\subsection{Methodology}
\label{app:experiments:methodology}

\paragraph{TF-IDF baseline.}
We extract word unigram TF-IDF features and tune $\min\_df \in \{1,2,3\}$ and $\max\_df \in \{0.9,0.95,1.0\}$. We use an L2-regularized logistic regression classifier with class-balanced weights. We perform a grid search over $C \in \{0.25, \allowbreak 0.5, \allowbreak 1, \allowbreak 2, \allowbreak 3, \allowbreak 5, \allowbreak 10\}$ and select the best model based on development set macro-F1.

For CLARITY, the best configuration is $\min\_df{=}2$, $\max\_df{=}0.95$, $C{=}0.5$ for clarity, and $\min\_df{=}2$, $\max\_df{=}0.9$, $C{=}2$ for evasion. For PolERo, the best parameters are $\min\_df{=}3$, $\max\_df{=}0.9$, $C{=}5$ for clarity and $\min\_df{=}1$, $\max\_df{=}0.9$, $C{=}2$ for evasion.

\subsubsection*{Encoder Fine-Tuning}

\paragraph{Input formatting.}
Encoder models use the input format ``\texttt{Question: \{q\}\textbackslash nAnswer: \{a\}}''. On CLARITY, $q$ is the GPT-decomposed atomic \texttt{question} field paired with the \texttt{interview\_answer}. On PolERo, $q$ is the \texttt{interview\_question} field paired with the same \texttt{interview\_answer}, since the Romanian dataset already contains single-question inputs.

\paragraph{Single-head encoders.}
Each of the eight single-head encoder configurations is fine-tuned separately for each task for up to 20 epochs using AdamW, with a batch size of 64. The maximum sequence length is set to 512 for all models except ModernBERT-large, which uses a maximum sequence length of 8192. The best checkpoint is selected on an $85/15$ train/validation split of the official training set and evaluated once on the test set. Learning rates vary by architecture: $5 \times 10^{-6}$ for ModernBERT-large and ELECTRA-large; $1\!\times\!10^{-5}$ for RoBERTa-large and XLM-RoBERTa-large; and $2\!\times\!10^{-5}$ for BERT-large-cased, mBERT-base, \textbf{Ro}BERT-large, and Romanian BERT.

\paragraph{Multi-head architecture.}
We apply this architecture to three encoder configurations: RoBERTa-large, XLM-RoBERTa-large, and \textbf{Ro}BERT-large. All multi-head models use 7-fold cross-validation stratified by the clarity label. Models are trained for 20 epochs using AdamW with a learning rate of $1\!\times\!10^{-5}$ and a batch size of 8. We use early stopping based on the combined validation macro-F1 of both heads, with a patience of 5. At inference time, we ensemble all 7 folds by averaging their predicted class probabilities and selecting the most probable class for each classification head.

\subsubsection*{Large Language Models}

Open-source LLMs were run locally, while proprietary models were accessed via the OpenAI API platform\footnote{\url{https://platform.openai.com}}. All models are queried with default settings and reasoning configurations, using a restricted output schema. Zero-shot prompts include the task description, the full set of category names, and one-line definitions. Few-shot prompts add one labeled example per category. Prompts are run separately for the clarity and evasion tasks and use the dataset’s language.

\section{Multi-Head Architecture Details}
\label{app:multihead}

This appendix expands the multi-head architecture introduced in Section~\ref{sec:setup} and isolates the contribution of each component on the English dataset. All ablation numbers are 7-fold cross-validation means and standard deviations of validation macro-F1, computed on the training split with folds stratified by the clarity label. Test-set scores for this configuration are reported in Table~\ref{tab:full_results}.

\subsection{Hierarchical Input Processing}
\label{app:multihead:chunking}

Responses in both datasets are often longer than the 512-token limit of standard pretrained encoders. Because evasion cues can appear anywhere in a response, simple truncation may remove the relevant span. To avoid this, we segment each tokenized question-answer sequence $T$ into overlapping windows of length $L{=}512$ with stride $S{=}256$:
\begin{equation}
    C_k = T[kS : kS + L] \quad \text{for } k=0,\ldots,M-1,
\end{equation}
where $M = \lceil \max(|T|-L,0)/S \rceil + 1$. The final chunk extends to $|T|$ and is zero-padded if shorter. The window size $L$ is fixed by the encoder's positional embedding capacity. The 50\% overlap ensures that every interior token is included in at least two consecutive windows, reducing the chance that important evasion cues are separated across chunks.

Each chunk is encoded independently by the shared encoder. We extract the hidden state at position 0 of each chunk, $h_k = H_k[0,:] \in \mathbb{R}^{d}$, and aggregate the $M$ chunk vectors via element-wise max-pooling:
\begin{equation}
    v_j = \max_{k=0}^{M-1} h_{k,j} \quad \text{for } j=1,\ldots,d.
\end{equation}
The position-0 token corresponds to the encoder's start-of-sequence token only for the first chunk; for subsequent chunks, it is the first content token of the window. We use this position uniformly because the encoder distributes contextual information across all tokens.

The pooled vector $v$ is passed through dropout ($p{=}0.1$) and fed into two linear classification heads:
\begin{align}
\hat{y}_c &= \mathrm{softmax}(W_c \cdot \mathrm{Dropout}(v) + b_c), \\
\hat{y}_e &= \mathrm{softmax}(W_e \cdot \mathrm{Dropout}(v) + b_e),
\end{align}
with $W_c \in \mathbb{R}^{3 \times d}$ and $W_e \in \mathbb{R}^{9 \times d}$. The training objective is the unweighted sum of two cross-entropy losses, $\mathcal{L} = \mathcal{L}_c + \mathcal{L}_e$. At inference time, the 7 models are ensembled by averaging predicted class probabilities and taking the argmax per head.

\subsection{Component Ablations}
\label{app:multihead:ablations}

The ablations below isolate the impact of each architectural choice over single-head encoder fine-tuning.

\paragraph{Pooling strategy.}
Table~\ref{tab:ablation_pooling} compares element-wise max-pooling with mean-pooling and a baseline that keeps only the first chunk representation. Max-pooling performs best. Mean-pooling performs worse than max-pooling, indicating that simple averaging is less effective for aggregating chunk representations. The first-chunk baseline misses any signal beyond the first 512 tokens, which affects $28.8\%$ of English responses.

\begin{table}[H]
\centering
\small
\begin{tabular}{lcc}
\toprule
\textbf{Pooling} & \textbf{Clarity} & \textbf{Evasion} \\
\midrule
First chunk only & $0.67 \pm 0.01$ & $0.42 \pm 0.01$ \\
Mean-pooling     & $0.68 \pm 0.02$ & $0.43 \pm 0.02$ \\
Max-pooling      & $\mathbf{0.70 \pm 0.02}$ & $\mathbf{0.45 \pm 0.02}$ \\
\bottomrule
\end{tabular}
\caption{Chunk aggregation strategy. 7-fold CV validation macro-F1 (mean $\pm$ std).}
\label{tab:ablation_pooling}
\end{table}

\paragraph{Multi-task versus single-task.}
Table~\ref{tab:ablation_mt} shows that joint training matches single-task performance on clarity and improves single-task evasion by $+0.03$ macro-F1. The coarser clarity supervision acts as a regularizer for rare evasion categories: each evasion label maps deterministically to a clarity label, so gradients from $\mathcal{L}_c$ help stabilize representations of the parent class, even when the evasion head provides limited signal.

\begin{table}[H]
\centering
\small
\begin{tabular}{lcc}
\toprule
\textbf{Objective} & \textbf{Clarity} & \textbf{Evasion} \\
\midrule
Single-task (clarity)  & $\mathbf{0.70 \pm 0.02}$ & -- \\
Single-task (evasion)  & -- & $0.42 \pm 0.01$ \\
Multi-task             & $0.70 \pm 0.02$ & $\mathbf{0.45 \pm 0.02}$ \\
\bottomrule
\end{tabular}
\caption{Effect of joint training. 7-fold CV validation macro-F1 (mean $\pm$ std).}
\label{tab:ablation_mt}
\end{table}

\paragraph{Ensemble size.}
Table~\ref{tab:ablation_ensemble} reports performance for $k \in \{3,5,7\}$ folds. Increasing $k$ reduces variance and increases the amount of training data per fold model. Improvements are consistent across both subtasks and larger for the more difficult evasion task.

\begin{table}[H]
\centering
\small
\begin{tabular}{lcc}
\toprule
\textbf{Folds} & \textbf{Clarity} & \textbf{Evasion} \\
\midrule
3 & $0.66 \pm 0.01$ & $0.42 \pm 0.02$ \\
5 & $0.68 \pm 0.02$ & $0.43 \pm 0.03$ \\
7 & $\mathbf{0.70 \pm 0.02}$ & $\mathbf{0.45 \pm 0.02}$ \\
\bottomrule
\end{tabular}
\caption{Effect of ensemble size. $k$-fold CV validation macro-F1 (mean $\pm$ std).}
\label{tab:ablation_ensemble}
\end{table}

\paragraph{Loss function.}
Table~\ref{tab:ablation_loss} compares unweighted cross-entropy with inverse-frequency class-weighted cross-entropy and focal loss ($\gamma{=}2$) \cite{8417976}. Neither alternative improves macro-F1. Class weighting increases minority-class recall (\textit{Clear Non-Reply} from 0.62 to 0.67, \textit{Partial/half-answer} from 0.00 to 0.10), but this is offset by lower precision on majority classes. Focal loss yields intermediate performance between the two. Most evasion errors occur between pragmatically similar classes (\textit{Implicit}, \textit{Deflection}, \textit{General}, \textit{Dodging}) that share surface features. Reweighting increases the training signal for rare classes but does not reduce overlap between similar evasion categories.

\begin{table}[H]
\centering
\resizebox{\columnwidth}{!}{%
\begin{tabular}{lcc}
\toprule
\textbf{Loss} & \textbf{Clarity} & \textbf{Evasion} \\
\midrule
Cross-entropy (unweighted) & $\mathbf{0.70 \pm 0.02}$ & $\mathbf{0.45 \pm 0.02}$ \\
Class-weighted CE          & $0.66 \pm 0.02$ & $0.41 \pm 0.02$ \\
Focal ($\gamma{=}2$)       & $0.69 \pm 0.02$ & $0.44 \pm 0.02$ \\
\bottomrule
\end{tabular}%
}
\caption{Effect of loss function. 7-fold CV validation macro-F1 (mean $\pm$ std).}
\label{tab:ablation_loss}
\end{table}

\subsection{Seed Variation}
\label{app:experiments:seeds}

We repeat the full multi-head pipeline for both monolingual configurations under three random seeds ($7$, $42$, $123$). The seed controls the cross-validation split, the per-fold training seed, and weight initialization. Table~\ref{tab:seed_variation} reports mean and standard deviation of macro-F1.

Cross-validation scores are stable: combined macro-F1 varies by $\pm.002$ on English and $\pm.008$ on Romanian, as each figure averages over seven folds. Test-set scores vary more, as they come from a single evaluation: clarity is $.676_{\pm.019}$ on English and $.743_{\pm.023}$ on Romanian, evasion $.535_{\pm.024}$ and $.558_{\pm.037}$. This variation is smaller than the differences we report between model families, so the effects cannot be attributed to seed choice.

\begin{table}[H]
\centering
\small
\renewcommand{\arraystretch}{1.15}
\begin{tabular*}{\columnwidth}{@{\extracolsep{\fill}}l rr@{}}
\toprule
\textbf{Metric} & \textbf{English} & \textbf{Romanian} \\
\midrule
\multicolumn{3}{@{}l}{\textit{7-fold cross-validation (validation folds)}} \\
Clarity          & $.700_{\pm.001}$ & $.676_{\pm.006}$ \\
Evasion          & $.450_{\pm.004}$ & $.388_{\pm.011}$ \\
Combined         & $.575_{\pm.002}$ & $.532_{\pm.008}$ \\
\midrule
\multicolumn{3}{@{}l}{\textit{Test set (7-fold ensemble)}} \\
Clarity          & $.676_{\pm.019}$ & $.743_{\pm.023}$ \\
Evasion (any-of) & $.535_{\pm.024}$ & $.558_{\pm.037}$ \\
Mapped-clarity   & $.682_{\pm.021}$ & $.739_{\pm.027}$ \\
\bottomrule
\end{tabular*}
\caption{Seed variation of the multi-head architecture: macro-F1 mean $\pm$ standard deviation over three seeds ($7$, $42$, $123$), with RoBERTa-large as the English backbone and \textbf{Ro}BERT-large as the Romanian one.}
\label{tab:seed_variation}
\end{table}

\section{Extended Model Comparison and Error Analysis}
\label{app:experiments:comparison}

Table~\ref{tab:full_results} presents the complete test-set macro-F1 results for all baselines, architectural variants, cross-lingual transfer setups, and LLM prompting strategies discussed in this work.

\subsection{Traditional classifier}
The TF-IDF logistic regression baseline achieves similar macro-F1 on both languages ($0.42$ on English, $0.45$ on Romanian), but its error patterns show which parts of the task rely on lexical cues. The model correctly identifies most \textit{Clear Reply} instances in Romanian (64\% recall), but performs much worse in English (33\% recall). 

In contrast, the \textit{Ambivalent} class is often confused with the other two labels. In Romanian, this misclassification goes both ways: 48\% of true \textit{Ambivalent} instances are predicted as \textit{Clear Reply}, while 33\% of true \textit{Clear Reply} instances are predicted as \textit{Ambivalent}. This suggests that any answer that reuses question vocabulary is often treated as a reply by the baseline, even when it fails to provide a concrete answer.

\textit{Clear Non-Reply} performs worst, with an F1 of 0.25 on English and 0.26 on Romanian (and recall as low as 24\%). As reported in Appendix~\ref{sec:eda:length}, these are typically very short non-replies (e.g., \textit{Clarification}, \textit{Declining to answer}). Because this classifier relies on term frequency, these brief answers generate weak feature signals, causing the model to underpredict these minority classes.

The slightly higher performance on Romanian is consistent with the higher share of \textit{Explicit} replies in its training distribution (Section~\ref{sec:eda:labels}). More cases can be solved through lexical overlap.

\subsection{Encoder Models}

\paragraph{Best single-head models.}
Among single-head encoders, RoBERTa-large is the strongest English model on evasion. 

\textbf{Ro}BERT-large is the strongest model on PolERo. It is pretrained on Romanian text and uses a tokenizer trained on Romanian data, unlike the shared multilingual vocabulary used by XLM-RoBERTa. Although XLM-RoBERTa-large is the strongest multilingual model among those evaluated and is competitive with monolingual encoders on Romanian clarity, it underperforms on the fine-grained evasion task.

\paragraph{Failure modes.}
ELECTRA-large and ModernBERT-large are the two main failure cases. ELECTRA uses a replaced-token-detection (RTD) objective, which trains the model at the token level to distinguish real from replaced tokens. On small and imbalanced datasets like those tested, we hypothesize that RTD may make the model more susceptible to overfitting on frequent lexical cues rather than learning broader evasion features. As a result, on the Romanian dataset, the model assigns all inputs to the \textit{Explicit} class.

ModernBERT-large shows the same issue, though less severely. Its local-global attention mechanism improves efficiency for longer input sequences compared to standard BERT encoders, which may offer limited benefit for short question-answer inputs. In our experiments, three rare categories (\textit{Claims ignorance}, \textit{Partial/half-answer}, \textit{Deflection}) receive zero recall on Romanian, and the mapped clarity predictions inherit the same majority-class bias.

\paragraph{Multi-head architecture.}
Multi-head models consistently outperform single-head models across all reported metrics. The strongest performance is observed for RoBERTa-large (English) and \textbf{Ro}BERT-large (Romanian), both of which are pretrained on in-language corpora and appear to benefit most from the joint signal in our experiments. XLM-RoBERTa-large shows larger relative gains, but still underperforms monolingual encoders, suggesting that multilingual representations may not fully capture fine-grained signals within a given language.

\begin{figure}[ht]
    \centering
    \includegraphics[width=\linewidth]{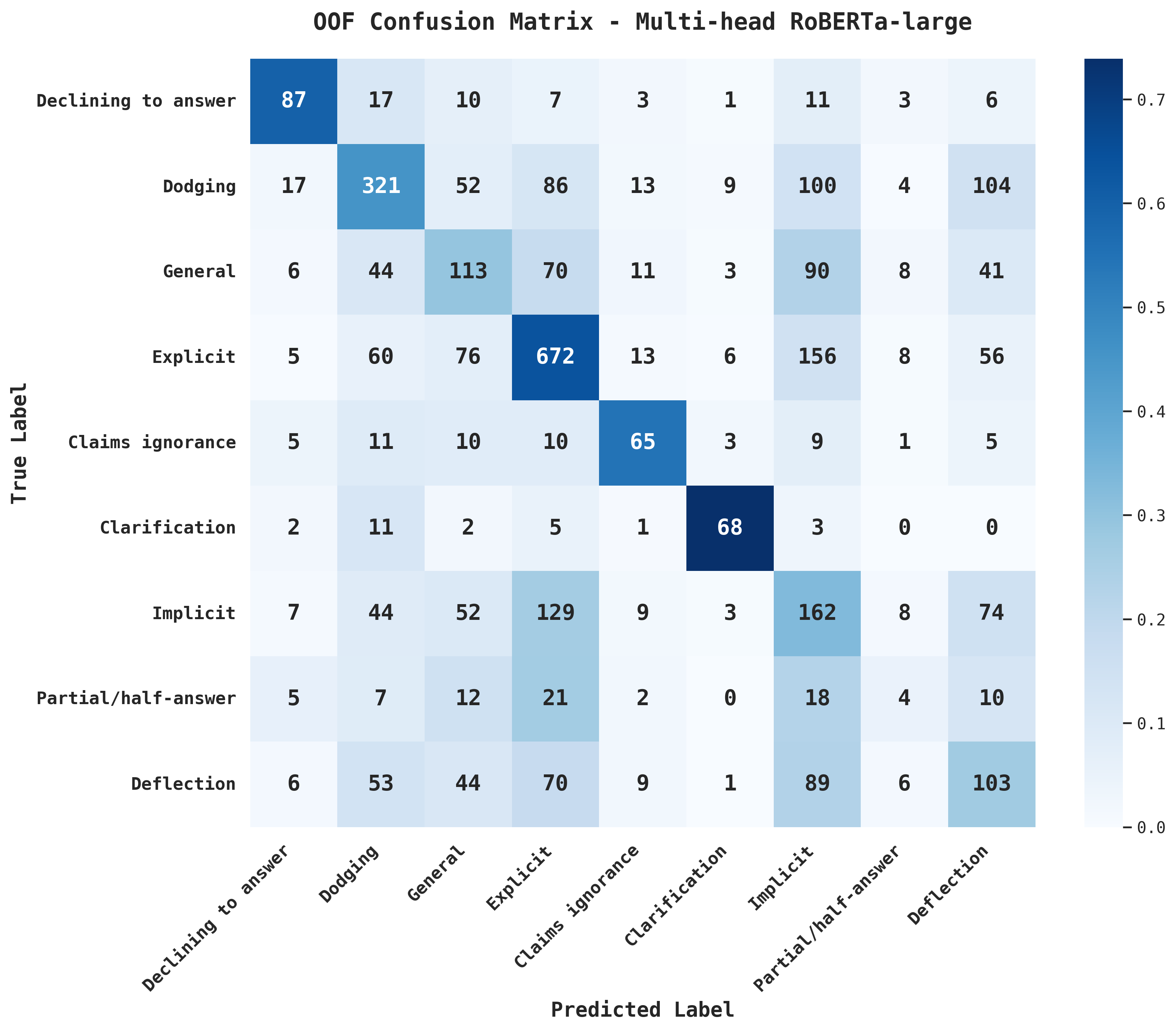} 
    \caption{Out-of-fold (OOF) confusion matrix for the 9-class evasion task on CLARITY (EN) using multi-head RoBERTa-large.}
    \label{fig:cm_en}
\end{figure}

\begin{figure}[ht]
    \centering
    \includegraphics[width=\linewidth]{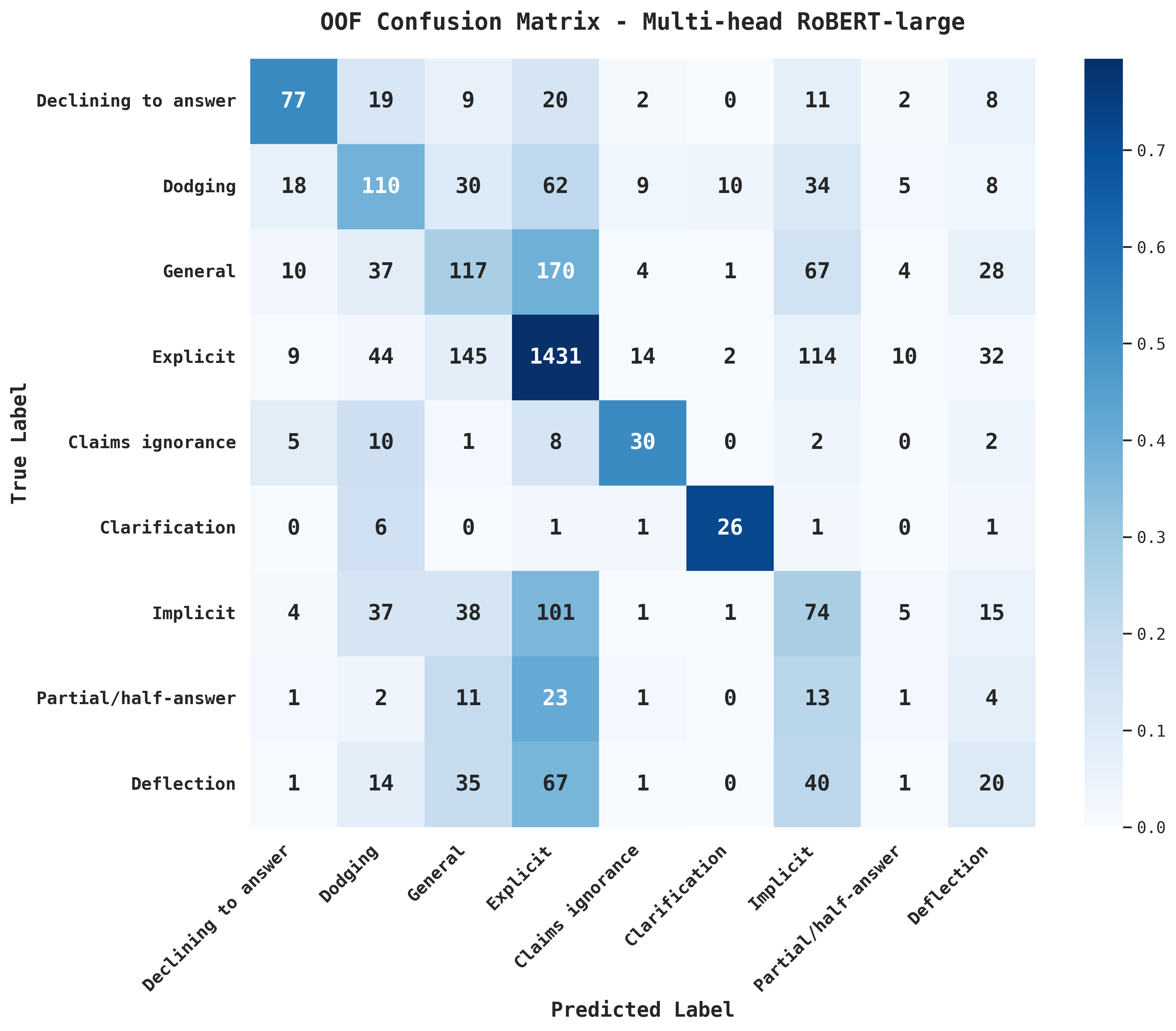} 
    \caption{Out-of-fold (OOF) confusion matrix for the 9-class evasion task on PolERo (RO) using multi-head \textbf{Ro}BERT-large.}
    \label{fig:cm_ro}
\end{figure}
\paragraph{Error Analysis.}

While the boundary between \textit{Clear Reply} and \textit{Ambivalent} remains the largest source of errors, the out-of-fold confusion matrices for our strongest models (Figure~\ref{fig:cm_en}, Figure~\ref{fig:cm_ro}) show that the fine-grained errors between evasion strategies concentrate along two main semantic axes. 

The most prominent axis of confusion is \textit{Implicit}--\textit{Explicit}. Both represent valid, on-topic answers. The difference depends on whether the requested information is stated directly or must be inferred from context. In English, this axis accounts for $285$ errors, and it remains a major source of error in Romanian with $215$ misclassifications.

The second major axis is \textit{General}-\textit{Explicit}. Both classes provide relevant information in response to the requested information. The distinction depends on whether the response directly addresses the specific details of the question or remains more general. This axis accounts for $146$ errors in English, but increases substantially in Romanian, becoming the largest source of error with $315$ misclassifications.

Another distinct source of errors is \textit{Dodging}-\textit{Deflection}. Both involve topic shifts. The difference depends on whether the model first attempts to answer the question before shifting topic or shifts topic without addressing the question. In English, this axis accounts for $157$ errors, although decreasing to $22$ errors in Romanian.

\textit{Partial/half-answer} is the most difficult category for all encoders, with an F1 of $0.066$ on English and $0.024$ on Romanian. Correctly identifying it requires determining whether only part of a multi-part question is answered while another part is ignored, a relational judgment that standard encoder representations do not capture well.

\begin{table*}[!htb]
\centering
\small
\begin{tabular}{l ccc ccc ccc}
\toprule
& \multicolumn{3}{c}{\textbf{Single-head}} & \multicolumn{3}{c}{\textbf{Multi-head}} & \multicolumn{3}{c}{$\Delta$ \textbf{(MH $-$ SH)}} \\
\cmidrule(lr){2-4}\cmidrule(lr){5-7}\cmidrule(lr){8-10}
\textbf{Configuration} & C & E & M & C & E & M & C & E & M \\
\midrule
EN $\rightarrow$ RO           & .506 & .282 & .512 & .635 & .399 & .655 & $+$.129 & $+$.117 & $+$.143 \\
EN2RO $\rightarrow$ RO        & .565 & .314 & .616 & .665 & .412 & .706 & $+$.100 & $+$.098 & $+$.090 \\
EN + RO $\rightarrow$ RO      & .695 & .403 & .686 & .723 & .570 & .717 & $+$.028 & $+$.167 & $+$.031 \\
EN $\rightarrow$ EN2RO        & .535 & .334 & .574 & .616 & .448 & .612 & $+$.081 & $+$.114 & $+$.038 \\
EN + RO $\rightarrow$ EN      & .628 & .459 & .617 & .679 & .483 & .642 & $+$.051 & $+$.024 & $+$.025 \\
EN + RO $\rightarrow$ EN + RO & .676 & .508 & .688 & .737 & .529 & .705 & $+$.061 & $+$.021 & $+$.017 \\
\midrule
\textit{Mean gain} & & & & & & & $\mathbf{+.075}$ & $\mathbf{+.090}$ & $\mathbf{+.057}$ \\
\bottomrule
\end{tabular}
\caption{Single-head vs.\ multi-head XLM-RoBERTa-large across cross-lingual configurations. \textbf{C}: clarity, \textbf{E}: evasion, \textbf{M}: mapped clarity; test-set macro-F1.}
\label{tab:sh_vs_mh}
\end{table*}

\paragraph{Error structure.}
We inspect where the errors of the strongest fine-tuned encoders happen relative to annotator agreement on the test split. For multi-head RoBERTa-large on the English evasion task, $43.7\%$ of errors occur on items with full $3/3$ annotator agreement and $50.4\%$ on items with $2/3$ agreement, while only $5.9\%$ involve perfect disagreement. The pattern is more evident on multi-head \textbf{Ro}BERT-large for Romanian, where $54.7\%$ of evasion errors and $56.6\%$ of clarity errors fall on items with unanimous labels. Errors occur on items with the cleanest annotation signal rather than on the genuinely ambiguous ones, largely because unanimous and high-agreement items make up the vast majority of the dataset. The higher apparent accuracy on the perfect-disagreement subset ($.758$ on English evasion vs.\ $.528$ on $3/3$ items) is an expected result of the evasion evaluation rule (Section \ref{app:experiments:evaluation}), which marks a prediction as correct if it matches any of the three distinct annotator labels.

\paragraph{Calibration.}
The difference between top-1 probability on correct and incorrect predictions is narrow across both tasks. The English RoBERTa-large multi-head averages $.605$ on correct evasion predictions and $.483$ on incorrect ones. The clarity head averages $.835$ vs.\ $.726$. The Romanian \textbf{Ro}BERT-large multi-head is more confident in both cases ($.923$ vs.\ $.822$ for clarity; $.859$ vs.\ $.637$ for evasion), consistent with the more imbalanced Romanian label distribution. Confident errors occur in both languages: the Romanian clarity head assigns probability $1.00$ to \textit{Clear Reply} on single-word responses unanimously labeled \textit{Ambivalent}. The English evasion head assigns probability above $.85$ to \textit{Explicit} on $3/3$-agreement \textit{Implicit} and \textit{Deflection} responses. Top-1 probability is therefore not a reliable rejection signal at deployment.

\paragraph{Single-head vs.\ multi-head.}
Table~\ref{tab:sh_vs_mh} isolates the architectural contribution from the training-data
configuration by running each cross-lingual setting with both a standard single-head encoder and the multi-head architecture, using XLM-RoBERTa-large as a common backbone with identical data, input format and evaluation protocol. The multi-head architecture outperforms the single-head baseline in every configuration and on every metric, by a mean of $+.075$ macro-F1 on clarity, $+.090$ on evasion and $+.057$ on mapped clarity. The gains are therefore attributable to the architecture rather than to the training-data configuration. The advantage is largest on the fine-grained evasion task, consistent with the joint objective stabilising rare categories. It is also largest where training data is scarcest: zero-shot transfer (EN $\rightarrow$ RO) gains $+.129$ on clarity, whereas the joint bilingual settings gain $+.028$ and $+.051$.

\subsection{LLM Prompting Analysis}
\label{app:llmpromptanalys}

The complete LLM results in \autoref{tab:full_results} show that GPT-5.4 achieves the best mapped scores on both languages and the strongest English evasion performance. Qwen3.6-35B-A3B performs best on Romanian evasion under few-shot prompting, while Gemma-4-31B-it is best on English direct clarity.

For both languages, the performance difference between the strongest LLMs and encoder ensembles is larger on the evasion task than on the clarity task. The clarity task mainly requires checking whether the answer matches the question, which strong encoders handle well. In contrast, the nine-class evasion task requires distinguishing between closely related rhetorical strategies and benefits from broader discourse and pragmatic knowledge.

Multi-head fine-tuning reduces this difference, but does not eliminate it. Performance on the ambivalent evasion categories remains the main challenge for both model families.

\subsection{Machine Translation}

Translate-Train recovers part of the mapped clarity gap but yields only small improvements on the evasion task. Machine translation preserves enough propositional content to recover the coarse three-class distinction, but distorts pragmatic cues that separate ambivalent categories.

\textit{Deflection} and \textit{Dodging} are most affected, as both rely more on discourse structure than lexical cues. As a result, both show substantial recall drops under translation compared to in-language single-head models. On the Romanian test set, recall for \textit{Deflection} decreases from 41\% to 10\%, while \textit{Dodging} drops from 50\% to 30\%.

\subsection{Machine Translation Quality}
\label{app:mt_quality}
We evaluate the quality of the NLLB-200-distilled-600M translations used in the cross-lingual experiments with XCOMET-XL, a model that estimates translation quality without reference translations. Results are reported for both translation directions on the training and test splits.

For RO$\rightarrow$EN, translating the Romanian data resulted in mean QE scores of $0.650$ (train, $n{=}3{,}278$) and $0.655$ (test, $n{=}296$), with clear variation across categories. Formulaic categories translate well, including \textit{Clarification} ($0.781$ train, $0.941$ test) and \textit{Declining to answer} ($0.763$ train, $0.763$ test), which obtain the highest scores. Categories driven by pragmatics score lower. \textit{Deflection} has the lowest scores in both splits ($0.582$ train, $0.473$ test). This drop matches the recall degradation for \textit{Deflection} and \textit{Dodging} under MT augmentation. 

For EN$\rightarrow$RO, translating the English data resulted in mean QE scores of $0.218$ (train, $n{=}3{,}448$) and $0.171$ (test, $n{=}308$). Although these scores are notably low, a manual inspection of the translations shows that many spans flagged as errors are acceptable translations. Despite this noise, the relative ranking across categories is consistent with the RO$\rightarrow$EN direction: \textit{Deflection} and \textit{General} score lowest, while \textit{Clarification} and \textit{Claims ignorance} score highest.

\begin{table*}[ht]
\centering
\small
\renewcommand{\arraystretch}{1.15}
\begin{tabular}{@{}p{1.8cm}p{2.0cm}p{4.0cm}p{6.6cm}@{}}
\toprule
Clarity (L1) & Evasion (L2) & Description & Example (Romanian; English in italics) \\
\midrule
\textit{Clear Reply} & Explicit & The requested information is given directly, in the form requested, without ambiguity or required inference.
& \textbf{Î:} Când va fi gata autostrada? \newline \textbf{R:} Lucrările vor fi finalizate în luna martie 2025. \newline \textit{Q: When will the highway be ready? A: Works will be completed in March 2025.} \\
\midrule
\multirow{15}{*}{\textit{Ambivalent}}
& Implicit & The requested information is conveyed but not stated in the expected form; the listener must infer the conclusion.
& \textbf{Î:} Vă veți da demisia din funcția de ministru? \newline \textbf{R:} Nu mai pot rămâne o secundă în acest guvern. \newline \textit{Q: Will you resign as minister? A: I cannot stay in this government a second longer.} \\[2pt]
& General & The response stays on topic but remains too broad, abstract, or vague to deliver the requested specifics.
& \textbf{Î:} În ce lună va fi gata autostrada? \newline \textbf{R:} Dezvoltarea infrastructurii e o prioritate zero pentru noi și muncim zi de zi pe șantiere. \newline \textit{Q: In which month will the highway be ready? A: Infrastructure development is our top priority and we work every day on site.} \\[2pt]
& Partial/half-answer & Covers concretely only one component of the requested information and omits the rest.
& \textbf{Î:} Ce companii de stat vor fi privatizate anul acesta? \newline \textbf{R:} Vă pot spune cu certitudine că Poșta Română se află pe această listă. \newline \textit{Q: Which state-owned companies will be privatised this year? A: I can tell you with certainty that the Romanian Postal Service is on the list.} \\[2pt]
& Dodging & Does not engage with the substance of the question; the topic is neither acknowledged nor touched.
& \textbf{Î:} De ce ați pierdut alegerile locale? \newline \textbf{R:} Presa ar trebui să se ocupe de problemele reale ale cetățenilor. \newline \textit{Q: Why did you lose the local elections? A: The press should focus on citizens' real problems.} \\[2pt]
& Deflection & Acknowledges the topic of the question but shifts the argument and focus to something other than what is asked.
& \textbf{Î:} Câți bani ați alocat pentru acest program? \newline \textbf{R:} Programul este un proiect la care ținem foarte mult, însă marea realizare de anul acesta este că am reușit să modernizăm peste 50 de școli. \newline \textit{Q: How much money did you allocate to this programme? A: The programme is a project we care deeply about, but this year's big achievement is that we modernised over 50 schools.} \\
\midrule
\multirow{10}{*}{\shortstack[l]{\textit{Clear} \\ \textit{Non-Reply}}}
& Declining to answer & A verbal, assumed and explicit refusal to provide the requested information.
& \textbf{Î:} Când veți publica raportul financiar? \newline \textbf{R:} Nu voi face niciun comentariu pe acest subiect până la finalizarea auditului. \newline \textit{Q: When will you publish the financial report? A: I will not comment on this matter until the audit is complete.} \\[2pt]
& Claims ignorance & The speaker explicitly admits not holding the requested information at the moment.
& \textbf{Î:} La ce dată exactă a ordonat guvernul reamenajarea navei HMAS Kanimbla? \newline \textbf{R:} Nu știu acea dată. Voi afla și voi informa Camera. \newline \textit{Q: On what exact date did the government order the refit of HMAS Kanimbla? A: I do not know that date. I will find out and inform the Chamber.} \\[2pt]
& Clarification & Does not provide the requested information and instead asks for clarification of what the question refers to.
& \textbf{Î:} A fost decizia dumneavoastră să eliberați fondul? \newline \textbf{R:} Vă referiți la fondul public sau cel privat? \newline \textit{Q: Was it your decision to release the fund? A: Do you mean the public fund or the private one?} \\
\bottomrule
\end{tabular}
\caption{PolERo taxonomy. Descriptions and examples are taken from the Romanian annotation guidelines used by annotators; English translations are provided in italics for readability.}
\label{tab:taxonomy_full}
\end{table*}

\begin{table*}[ht]
\centering
\small
\setlength{\tabcolsep}{5.5pt}
\begin{tabular}{l ccc ccc}
\toprule
& \multicolumn{3}{c}{\textbf{CLARITY (EN, $n{=}308$)}} & \multicolumn{3}{c}{\textbf{PolERo (RO, $n{=}296$)}} \\
\cmidrule(lr){2-4}\cmidrule(lr){5-7}
\textbf{Model} & Clarity & Evasion & Mapped & Clarity & Evasion & Mapped \\
\midrule
TF-IDF + Logistic Regression & .422 & .301 & .441 & .453 & .312 & .472 \\
\midrule
\multicolumn{7}{l}{\textit{Single-head encoder fine-tuning}} \\
\quad mBERT-base                   & .529 & .347 & .517 & .601 & .328 & .502 \\
\quad BERT-large-cased             & .587 & .361 & .581 & .475 & .282 & .501 \\
\quad ELECTRA-large                & .607 & .231 & .524 & .500 & .084 & .241 \\
\quad ModernBERT-large             & .575 & .360 & .524 & .440 & .246 & .443 \\
\quad RoBERTa-large                & .580 & .481 & .656 & .646 & .428 & .613 \\
\quad XLM-RoBERTa-large            & .579 & .419 & .564 & .680 & .365 & .686 \\
\quad Romanian BERT (RO)           & --   & --   & --   & .684 & .470 & .634 \\
\quad RoBERT-large (RO)            & --   & --   & --   & .688 & .504 & .701 \\
\midrule
\multicolumn{7}{l}{\textit{Multi-head encoder fine-tuning}} \\
\quad RoBERTa-large                & \underline{.713} & .568 & .731 & .696 & .501 & .652 \\
\quad XLM-RoBERTa-large            & .665 & .475 & .644 & .694 & .508 & .706 \\
\quad RoBERT-large (RO)            & --   & --   & --   & .729 & .545 & \textbf{.763} \\
\midrule
\multicolumn{7}{l}{\textit{Cross-lingual transfer (XLM-RoBERTa-large, multi-head)}} \\
\quad EN $\rightarrow$ RO                                    & --   & --   & --   & .635 & .399 & .655 \\
\quad EN2RO $\rightarrow$ RO                                 & --   & --   & --   & .665 & .412 & .706 \\
\quad RO $\rightarrow$ RO2EN                                 & --   & --   & --   & .681 & .494 & .702 \\
\quad RO + RO2EN $\rightarrow$ RO                            & --   & --   & --   & .708 & .492 & .681 \\
\quad RO + EN2RO $\rightarrow$ RO  & --   & --   & --   & .718 & .489 & .684 \\
\quad EN + RO $\rightarrow$ RO                               & --   & --   & --   & .723 & .570 & .717 \\
\quad RO $\rightarrow$ EN                                    & .551 & .398 & .527 & --   & --   & --   \\
\quad RO2EN $\rightarrow$ EN                                 & .477 & .358 & .439 & --   & --   & --   \\
\quad EN $\rightarrow$ EN2RO                                 & .616 & .448 & .612 & --   & --   & --   \\
\quad EN + EN2RO $\rightarrow$ EN                            & .621 & .460 & .647 & --   & --   & --   \\
\quad EN + RO2EN $\rightarrow$ EN  & .694 & .477 & .602 & --   & --   & --   \\
\quad EN + RO $\rightarrow$ EN                               & .679 & .483 & .642 & --   & --   & --   \\
\quad EN + RO $\rightarrow$ EN + RO \textsuperscript{\textdagger} & \multicolumn{6}{c}{Clarity .737 \quad Evasion .529 \quad Mapped .705} \\
\midrule
\multicolumn{7}{l}{\textit{Cross-lingual transfer (RoBERTa-large, multi-head)}} \\
\quad EN $\rightarrow$ RO                                    & --   & --   & --   & .624 & .409 & .639 \\
\quad EN2RO $\rightarrow$ RO & --   & --   & --   & .594 & .421 & .584 \\
\quad RO $\rightarrow$ RO2EN                                 & --   & --   & --   & .625 & .427 & .635 \\
\quad RO + RO2EN $\rightarrow$ RO                            & --   & --   & --   & .664 & .512 & .652 \\
\quad RO + EN2RO $\rightarrow$ RO & --   & --   & --   & .677 & .491 & .674 \\
\quad EN + RO $\rightarrow$ RO                               & --   & --   & --   & .712 & .463 & .681 \\
\quad RO $\rightarrow$ EN                                    & .572 & .337 & .499 & --   & --   & --   \\
\quad RO2EN $\rightarrow$ EN & .590 & .409 & .543 & --   & --   & --   \\
\quad EN $\rightarrow$ EN2RO                                 & .479 & .357 & .502 & --   & --   & --   \\
\quad EN + EN2RO $\rightarrow$ EN                            & .672 & .513 & .654 & --   & --   & --   \\
\quad EN + RO2EN $\rightarrow$ EN & .675 & .505 & .641 & --   & --   & --   \\
\quad EN + RO $\rightarrow$ EN                               & .671 & .503 & .683 & --   & --   & --   \\
\quad EN + RO $\rightarrow$ EN + RO \textsuperscript{\textdagger} & \multicolumn{6}{c}{Clarity .715 \quad Evasion .528 \quad Mapped .718} \\
\midrule
\multicolumn{7}{l}{\textit{LLM prompting (no fine-tuning)}} \\
\quad Llama-3.3-70B (zero-shot)        & .474 & .377 & .624 & .479 & .354 & .446 \\
\quad Llama-3.3-70B (few-shot)         & .535 & .382 & .564 & .555 & .312 & .411 \\
\quad Gemma-4-31B-it (zero-shot)       & .655 & .546 & .619 & .657 & .560 & .713 \\
\quad Gemma-4-31B-it (few-shot)        & \textbf{.721} & .583 & .662 & .672 & .619 & .738 \\
\quad GPT-OSS-120B (zero-shot)         & .548 & .559 & .683 & .575 & .544 & .740 \\
\quad GPT-OSS-120B (few-shot)          & .542 & .554 & .729 & .553 & .580 & .729 \\
\quad DeepSeek-V4-Pro (zero-shot)      & .599 & .512 & .693 & .604 & .556 & .696 \\
\quad DeepSeek-V4-Pro (few-shot)       & .601 & .545 & .670 & .627 & .594 & .751 \\
\quad Qwen3.6-35B-A3B (zero-shot)      & .673 & .536 & .719 & .684 & .631 & .738 \\
\quad Qwen3.6-35B-A3B (few-shot)       & .669 & .544 & \underline{.747} & \underline{.746} & \textbf{.682} & \underline{.760} \\
\quad GPT-5.4 (zero-shot)              & .687 & \underline{.613} & .729 & .687 & .649 & .744 \\
\quad GPT-5.4 (few-shot)               & .703 & \textbf{.626} & \textbf{.761} & \textbf{.763} & \underline{.666} & \textbf{.763} \\
\bottomrule
\end{tabular}
\caption{Full test-set macro-F1 across all models. \textbf{Clarity}: direct 3-class prediction. \textbf{Evasion}: direct 9-class prediction. \textbf{Mapped}: 3-class clarity prediction obtained by mapping the predicted 9-class evasion label upward through the taxonomy. For cross-lingual setups, \textit{EN2RO} denotes English data machine-translated into Romanian, and \textit{RO2EN} denotes Romanian data machine-translated into English. Best score per column is \textbf{bolded}; second-best is \underline{underlined}. \\
\textsuperscript{\textdagger} Evaluated on the combined EN+RO test set ($n{=}604$).}
\label{tab:full_results}
\end{table*}

\end{document}